%% file: neurips2021.tex
\documentclass{article}

   \PassOptionsToPackage{numbers, compress}{natbib}

\usepackage[preprint]{neurips_2021}

\usepackage[utf8]{inputenc} 
\usepackage[T1]{fontenc}  
\usepackage{hyperref}    
\usepackage{url}      
\usepackage{booktabs}    
\usepackage{amsfonts}    
\usepackage{nicefrac}    
\usepackage{microtype}   
\usepackage{xcolor}     
\usepackage{amssymb}
\usepackage{amsthm}
\usepackage{amsbsy}
\usepackage{amsmath}
\usepackage{subfig}
\usepackage{wrapfig}

\DeclareMathOperator*{\argmin}{arg\,min}
\usepackage{soul}
\setstcolor{red}

\newtheorem{assumption}{Assumption}
\newtheorem{theorem}{Theorem}
\newtheorem{definition}{Definition}
\newtheorem{proposition}{Proposition}
\newtheorem{assumption*}{Assumption}
\usepackage{bbm}

\usepackage{relsize}

\usepackage{color}
\usepackage[colorinlistoftodos]{todonotes}

\newcommand{\OpenQuotes}{``}
\newcommand{\CloseQuotes}{''}
\newcommand{\Quotes}[1]{{\OpenQuotes}#1{\CloseQuotes}}

\usepackage{cleveref}

\title{Sparsifying Binary Networks}

\author{Riccardo Schiavone \& Maria A. Zuluaga\\
	Data Science Department\\
	EURECOM\\
	Biot, France\\
	\texttt{\{riccardo.schiavone,maria.zuluaga\}@eurecom.fr} \\
}

\begin{document}

\maketitle

\begin{abstract}
Binary neural networks (BNNs) have demonstrated their ability to solve complex tasks with comparable accuracy as full-precision deep neural networks (DNNs), while also reducing computational power and storage requirements and increasing the processing speed. These properties make them an attractive alternative for the development and deployment of DNN-based applications in Internet-of-Things (IoT) devices. Despite the recent improvements, they suffer from a fixed and limited compression factor that may result insufficient for certain devices with very limited resources.
In this work, we propose sparse binary neural networks (SBNNs), a novel model and training scheme which introduces sparsity in BNNs and a new quantization function for binarizing the network's weights. The proposed SBNN is able to achieve high compression factors and it reduces the number of operations and parameters at inference time. We also provide tools to assist the SBNN design, while respecting hardware resource constraints. We study the generalization properties of our method for different compression factors through a set of experiments on linear and convolutional networks on three datasets. Our experiments confirm that SBNNs can achieve high compression rates, without compromising generalization, while further reducing the operations of BNNs, making SBNNs a viable option for deploying DNNs in cheap, low-cost, limited-resources IoT devices and sensors.

\end{abstract}

\section{Introduction}
	The term Internet-of-Things (IoT) became notable in the late 2000s under the idea of enabling internet access to electrical and electronic devices \cite{Miraz2015}, thus allowing these to collect and exchange data. The increasing number of connected IoT devices, which now surpasses the number of humans connected to the internet~\cite{evans2011}, has led to a sensors-rich world, capable of addressing a various number of real-time applications (e.g. security systems, healthcare monitoring, environmental meters, autonomous vehicles among others), where both accuracy and computational time are crucial~\cite{iot_application}.
	
Deep neural networks (DNNs) have demonstrated their ability to process large complex heterogeneous data and to extract patterns needed to take autonomous decisions with high reliability~\cite{bengioDNN}, reaching and surpassing state-of-the-art results for several tasks~\cite{natural_language_processing,speechrecognition,alexnet}. As such, DNNs have the potential of enabling a myriad of new IoT applications. 
However, DNNs are known for being resource-greedy, 
 in terms of required computational power, memory, and energy consumption~\cite{DNNanalysis}, whereas most IoT devices are characterized by limited resources. Indeed, they usually have limited processing power, small storage capabilities, they are not GPU-enabled and they are powered with batteries of limited capacity, which are expected to last over 10 years without being replaced or recharged \cite{GSMA_LTE_IoT}.
%
	These constraints represent an important bottleneck towards the deployment of DNNs in IoT applications~\cite{yao2018}.
	
	
	
	
Network sparsification, quantization and compression are a promising way to reduce DNNs' resource requirements and 
enable their usage in low power IoT devices. Under the principle that deep models contain optimal sub-networks that can perform the same task of their related super-network with less memory and reduced computational burden~\cite{denil2013predicting, winning_tickets}, pruning and quantization have been explored to extract these "lighter" sub-networks. Pruning addresses network sparsification by removing parameters and shared connections within the DNN for compression~\cite{han2015}. Sparsification reduces memory and power consumption, but the savings are limited by the need for floating-point operations over the full precision network parameters. Quantization addresses this by replacing floating-point operations and values with fixed-point ones through parameter discretization. A recent and notable example are binary neural networks (BNNs) \cite{Courbariaux2016}. These quantized networks, with weights and activation functions using one bit, avoid multiplication operations in the forward propagation, which are computationally expensive, replacing them with low-cost bitwise ones. The resulting sub-networks are faster at inference and achieve better compression, without compromising accuracy of complex learning tasks \citep{liu2020reactnet}. The main drawback of these techniques is their fixed and limited compression factor w.r.t. full-precision DNNs, 
which may result insufficient for certain limited size and low power embedded devices \citep{lin2020mcunet}.

In this work, we aim to address this limitation by proposing a novel quantized network, denoted as sparse binary neural network (SBNN), which shares the advantages of BNNs, as it performs quantization using one bit, but differently from these it introduces sparsity. Our main contributions are the following: 1) We formulate a novel binary network architecture, which accounts for sparsity by using 0/1 weights. 2) We present a simple yet effective SBNN training algorithm that builds on top of any state-of-the-art BNN training scheme, and we show an efficient implementation of it, which reduces the number of bitwise operations in the network proportionally to the network's sparsity. 3) We propose two theoretical compression rate bounds estimates of a SBNN as tools to assist the network's design by considering hardware memory resource constraints. 4) We evaluate the proposed SBNN on three benchmarks and find that our network achieves high compression rates without significantly compromising its performance accuracy. 
		
The remaining parts of this work are organized as follows. Section~\ref{sec:related} discusses previous related works. The core of our contribution is described in Section~\ref{sec:method}. In Section~\ref{sec:results}, we study the properties of the proposed method and assess its performance, in terms of accuracy and compression results, through a set of experiments using MNIST, CIFAR-10 and CIFAR-100 datasets. Finally, a discussion on the results and main conclusions are drawn in Section \ref{sec:conclusions}.
	
\section{Related Work}\label{sec:related}
The literature on network compression and speed-up is rich and includes different techniques, such as knowledge distillation~\cite{ba2014,Chen2017,hinton2015}, compact network design~\cite{Howard2017,Ma2018,Sandler2018,Szegedy2015}, quantization~\cite{compress_prune,courbariaux2017, yang2019quantization,Zhang2018} and sparsification through pruning~\cite{Bello1992,han2015,louizos2017,srivastava2014}. This work relies on the latter two techniques, which we review in the following for quantized networks. A broad review can be found in ~\cite{sze2017}. The section concludes by discussing BNNs training strategies.
	
\paragraph{Network Sparsification.}
The concept of sparsity has been well studied beyond quantized neural networks as it reduces a network's computational and storage requirements and it prevents overfitting. Methods to achieve sparsity either explicitly induce it during learning through regularization (e.g. $L_{0}$ \cite{louizos2017} or $L_{1}$ \cite{han2015} regularization), or do it incrementally by gradually augmenting small networks \cite{Bello1992}; or by post hoc pruning \cite{Gomez2019,sparse_using_binary,srivastava2014}.
	In the context of quantized networks, \citet{compress_prune} proposed magnitude-pruning of nearly-zero parameters from a trained full-precision dense model followed by a quantization step of the remaining weights. The method achieves high compression rates of $\sim$35-49$\times$ and inference speed-up on well-known DNN topologies, without incurring in accuracy losses. However the train-pruning stage is time-consuming and the achieved speed-up is relatively limited and model-dependent. \citet{tung2018clipq_compression} tried to optimize this scheme \cite{compress_prune}, reporting an improvement in accuracy, but at the cost of obtaining a smaller compression factor.
	Ternary neural networks (TNNs)~\cite{ternary2014} naturally perform magnitude-based pruning thanks to their quantization function, which maps real-valued weights to $\lbrace-1,0,+1\rbrace$, although some works \cite{faraone2017ternarycompressing, marban2020ternarylearning} have achieved larger compression rates by explicitly inducing sparsity through regularization. Being a binary network, for a given sparsity level our SBNN can achieve higher compression rates than TNNs, since it uses only two symbols instead of three. In contrast to current BNN formulations, which do not address sparsity explicitly and improve compression only through quantization \cite{xnornet}, the proposed SBNN can be represented using 0/1 weights, where the 1-valued weights are controlled through a new regularization term.   
	
	

	\paragraph{Quantization.}
	Network quantization allows the use of fixed-point arithmetic and a smaller bit-width to represent network parameters w.r.t the full-precision counterpart. 
	Representing the values using only a finite set requires a quantization function that maps the original elements to the finite set. The quantization can be done after training the model, using parameter sharing techniques \cite{compress_prune}, or during training by quantizing the weights in the forward pass, as TNNs \cite{ternary2014}, BNNs \cite{Courbariaux2015} and other quantized networks \cite{courbariaux2017, yang2019quantization} do. Our work adopts the same strategy, while proposing a quantization function that maps weights to a binary domain different from the standard antipodal one used in most state-of-the-art BNNs.
	

\paragraph{Binary Network Training.} \citet{Courbariaux2016} proposed a simple BNN training algorithm that uses a quantization function to map real-valued weights and activations to the $\lbrace-1,+1\rbrace$ binary domain. Subsequent works have extended \cite{Courbariaux2016} aiming to improve BNNs accuracy. For instance, \citet{xnornet} introduced a scaling coefficient to enable quantization to any pair of antipodal weights, whereas \citet{liu2018birealnet} proposed a smoother gradient function for the quantization operation. However, \citet{bethge2019backtosimplicity} showed that many of these extensions are not that effective in practice and simpler strategies (e.g. \cite{Courbariaux2016}) should be favored. Being a binary network, our work builds upon existing BNN training algorithms and extends them to account for sparsity by introducing a new regularization term in the BNN's objective loss. In this sense, any BNN training algorithm can be used at the core SBNN training. 


	\section{Method}\label{sec:method}
	We achieve sparsification in our SBNN by introducing a new penalization term in the objective loss, which controls the sparsity of the network (\Cref{problem_formulation_sec}). Being a BNN, we use BNN's state-of-the-art training algorithms for SBNN training by adding the sparsity regularization term to the original BNN's objective loss and defining a function to map standard BNN weights to the SBNN weight's domain (\Cref{training_sec}). We describe the implementation details of the proposed SBNN in \Cref{implementation_sec} to illustrate their speed-up gains w.r.t BNNs. Finally, as it is crucial to consider hardware resource constraints in the design of light-weight networks, we present two mathematical tools to estimate theoretical compression rate bounds for a given SBNN and thus assist its design (\Cref{design_sec}). 

	
	\subsection{Problem Formulation}\label{problem_formulation_sec}
The training of a full-precision DNN with $L$ layers can be seen as a loss minimization problem:
%
\begin{equation}\label{eq:standard}
\displaystyle
\argmin_{\tilde{\textbf{w}}} \mathcal{L}(y,\hat{y})
\end{equation}
where $\mathcal{L}(\cdot)$ is a loss function between the true labels $y$ and the predicted values $\hat{y}=f(\textbf{x};\tilde{\textbf{w}})$, which are a function of the data input $\textbf{x}$ and the network's full precision weights $\smash{\tilde{\textbf{w}}=\lbrace \tilde{\textbf{w}}^{(\ell)} \rbrace}$. We denote $\smash{\tilde{\textbf{w}}^{(\ell)} \in \mathbb{R}^{N_{\ell}}}$ the weights of the $\ell^{th}$ layer, $\ell \in \lbrace 1,\ldots,L \rbrace$, and $N=\sum_{\ell} N_{\ell}$ the total number of weights in the DNN.

Let us define the domain $\Omega=\lbrace \alpha, \beta \rbrace$, $\alpha, \beta \in \mathbb{R}$, and $S_\beta = \lbrace w_i| w_i=\beta\rbrace$, the set of network's weights $w_i$ with value $\beta$. In practice, we define a different per-layer domain $\smash{\Omega^{(\ell)}=\lbrace \alpha^{(\ell)}}, \beta^{(\ell)} \rbrace$, which accounts to redefining $S_\beta$ as $\smash{S_{\beta^{(\ell)}} = \lbrace w_i| w_i^{(\ell)}=\beta^{(\ell)}\rbrace}$. For the sake of simplicity in the notation and without loss of generality, in the following we choose to drop the layer index $\ell$ of $\smash{S_{\beta^{(\ell)}}}$ and $\Omega^{(\ell)}$, in favor of the more general $S_\beta$, $\Omega$ and its constituting parameters $\alpha, \beta$.

A sparse binary neural network (SBNN) is a network with weights $\textbf{w}=\lbrace \smash{\textbf{w}^{(\ell)}} \rbrace, \smash{\textbf{w}^{(\ell)}} \in \smash{\Omega^{N_{\ell}}} \,\, \forall \, \ell$, such that $|S_\beta| \leq M < N$. The loss minimization problem from \Cref{eq:standard} becomes a mixed discrete-continuous constrained optimization problem: 
\begin{equation}\label{min_problem_alphabeta_eq}
\begin{aligned}
& \argmin_{\textbf{w}} 
& & \mathcal{L}(y,\hat{y}) \\
& \text{s.t.}
& & \mathbf{w}^{(\ell)} \in \Omega^{N_{\ell}} \,\, \forall\ \ell, \\
&&& |S_\beta| \leq M < N.
\end{aligned}
\end{equation}

We define a mapping fuction $r: \lbrace 0,1 \rbrace \longrightarrow \lbrace \alpha, \beta \rbrace$ to express the SBNNs' weights in terms of 0/1 weights:
\begin{equation}\label{alphabeta_to_zeroone_eq}
\begin{aligned}
\displaystyle
w_i = r(w'_i) = (w'_i + \alpha') \cdot \beta'
\end{aligned}
\end{equation}
$w'_i \in \lbrace 0,1 \rbrace$ and $\alpha', \beta' \in \mathbb{R}$.

As a result of \Cref{alphabeta_to_zeroone_eq}, the predicted values $\hat{y}$ can be expressed as a function of $\alpha'$ and $\beta'$, i.e. $\hat{y}=f(\mathbf{x};\mathbf{w}',\alpha',\beta')$, where $\textbf{w}'= \{\mathbf{w}'^{(\ell)}\}$ and $\mathbf{w}'^{(\ell)} \in \{0,1\}^{N_{\ell}}$. The set $S_\beta$ can also be redefined in terms of one-valued weights, i.e. $S_1 = \lbrace w'_i| w'_i=1\rbrace$. \Cref{min_problem_alphabeta_eq} can be reformulated as:
\begin{equation}\label{min_problem_zeroone_eq}
\begin{aligned}
\displaystyle
& \argmin_{\textbf{w}',\alpha',\beta'}
& & \mathcal{L}(y,\hat{y}) \\
& \text{s.t.}
& & \textbf{w}'^{(\ell)} \in \{0,1\}^{N_{\ell}} \,\, \forall\ \ell, \\
&&& |S_1| \leq M < N,
\end{aligned}
\end{equation}
where the parameter $M$ relates to the sparsity of the network. The lower the value of $M$, the higher the sparsity of the SBNN. 

The mixed optimization problem in \Cref{min_problem_zeroone_eq} can be simplified by relaxing the sparsity constraint through the introduction of a non-negative function $\displaystyle g(\cdot)$, which penalizes the weights when $|S_1| > M$:
\begin{equation}\label{min_problem_zeroone_loss_eq}
\begin{aligned}
\displaystyle
& \argmin_{\textbf{w}',\alpha',\beta'}
& & \mathcal{L}(y,\hat{y}) + \lambda g(\textbf{w}')\\
& \text{s.t.}
& & \textbf{w}'^{(\ell)} \in \{0,1\}^{N_{\ell}} \,\, \forall\ \ell, \\
\end{aligned}
\end{equation}
and $\lambda$ controls the influence of $g(\cdot)$. A simple, yet effective function $g(\textbf{w}')$ is the following one:
\begin{equation}\label{penalty_function_eq}
g(\textbf{w}') = \text{ReLU}\left( \frac{|S_1|}{N} - \frac{M}{N}\right),
\end{equation}
where $M/N$ represents the fraction of expected connections (EC).

\subsection{Network Training}\label{training_sec}
We propose a SBNN training algorithm that builds upon state-of-the-art BNN training algorithms~\citep{bethge2019backtosimplicity,Courbariaux2016,liu2020reactnet}, while introducing network sparsification. Given $\mathcal{L}_{\text{\tiny{BNN}}}$ the objective loss of a BNN training algorithm, the SBNN training scheme extends it in two ways. First, it includes the penalization term to control sparsity (\Cref{penalty_function_eq}) and, second, it defines a function to map standard BNN weights to SBNN weights.

\paragraph{Penalization Term.}To profit from a BNN training scheme, SBNN training requires to add the penalization term in \Cref{penalty_function_eq} to the BNN objective loss to account for sparsity, i.e.
\begin{equation}\label{loss_bnn_sbnn_eq}
  \mathcal{L}_{\text{\tiny{SBNN}}} = \mathcal{L}_{\text{\tiny{BNN}}} + \lambda \,g(\textbf{w}').
\end{equation}
However, BNN training algorithms work with network's weights taking antipodal values. This requires to adapt the penalization term, as defined in \Cref{penalty_function_eq}, to consider antipodal weights, rather than on the SBNN's 0/1 weights. We define here the $i^{th}$ BNN weight as $w''_i \in \lbrace -1,+1\rbrace$, although any pair of antipodal values can be used. The adapted penalization function $h$ is:
\begin{equation}\label{actual_ones_eq}
h(\textbf{w}'') = \text{ReLU}\left(\left(\dfrac{1}{2N}\sum_i w''_i +1\right)-\text{EC}\right), 
\end{equation}
thus redefining the SBBN's objective loss as $\mathcal{L}_{\text{\tiny{SBNN}}} = \mathcal{L}_{\text{\tiny{BNN}}} + \lambda \, h(\textbf{w}'')$.

During training, we modulate the contribution of the regularization term $h(\textbf{w}'')$ by imposing, at every training iteration, to be equal to a fraction of $\mathcal{L}_{\text{\tiny{SBNN}}}$, i.e.
\begin{equation}\label{gamma_eq}
  \gamma = \dfrac{\lambda\,h(\textbf{w}'')}{\mathcal{L}_{\text{\tiny{SBNN}}}}.
\end{equation}
The hyperparameter $\gamma$ is set to a fixed value over all the training process. Since $\mathcal{L}_{\text{\tiny{SBNN}}}$ changes at every iteration, this forces $\lambda$ to adapt, thus modulating the influence of $h(\textbf{w}'')$ proportionally to the changes in the loss. The lower $\gamma$ is set, the less influence $h(\textbf{w}'')$ has on the total loss. This means that network sparsification will be slower, but convergence will be achieved faster. On the opposite case (high $\gamma$), the training will favor sparsification.


\paragraph{Weight Mapping.} The BNN weights $w''_i \in \{-1,+1\}$ need to be mapped to SBNNs weights $w'_i \in \{0,1\}$. To this end, we define a function $t : \{-1,+1\} \rightarrow \{\alpha,\beta\}$, which allows to map antipodal BNN weights $\textbf{w}''$ back to $\textbf{w}$:
\begin{equation}
  w_i = t(w''_i) = w''_i\cdot \beta'' + \alpha'', 
\end{equation}
where $\alpha'', \beta''$ are two learned parameters during training. After training, the final weights $\mathbf{w}$ are mapped to the $\{0,1\}$ domain according to \Cref{alphabeta_to_zeroone_eq}. We shall recall that for the sake of simplicity in the notation we have dropped the layer index in the formulations. However, in practice, the set of learned parameters differ on a per layer basis, i.e. $\{\alpha''^{(\ell)},\beta''^{(\ell)}\}$. 

\begin{figure}[t]
 \centering		 \includegraphics[width=\textwidth]{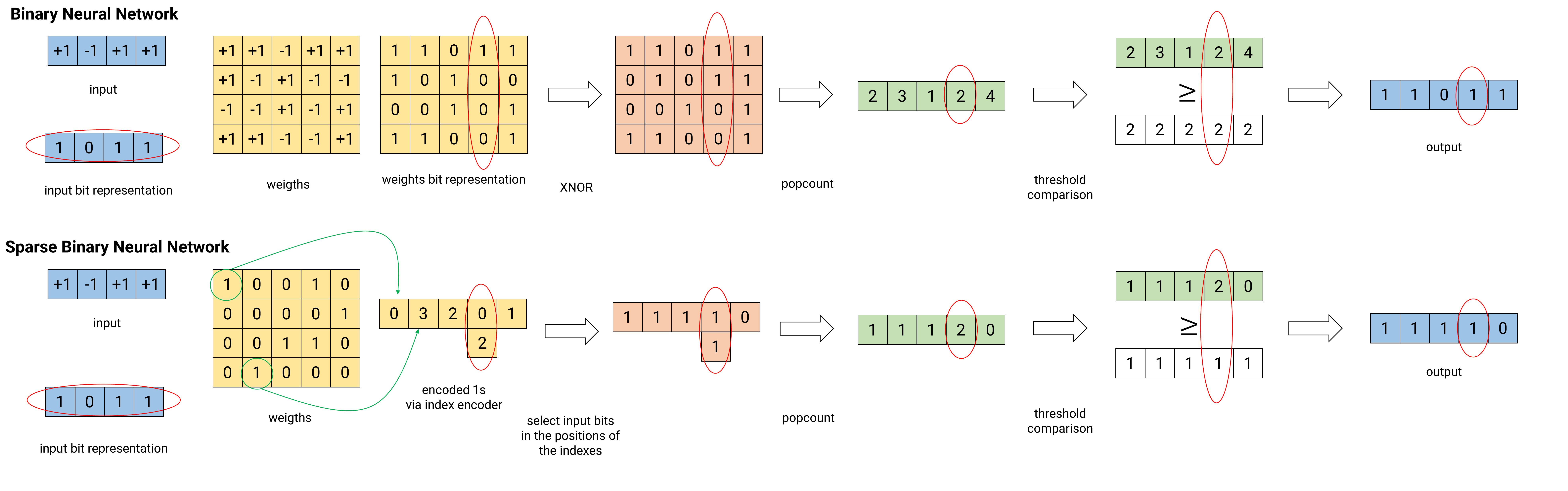}
					
		\caption{Example of implementation of $\mathbf{z}'=\mathbf{w'}\,\mathbf{x}$ operation in a linear layer of Binary Neural Networks and Sparse Binary Neural Networks. The red ellipses show the involved elements in the computation of the circled bit of the output vector on the right, while the green arrows give examples of the index encoding.}

		\label{implementation_fig}
	\end{figure}
	
\subsection{Implementation}\label{implementation_sec}
We illustrate the speed-up gains of the proposed SBNN through its efficient implementation using linear layers in the backbone architecture. Its extension to convolutional layers is straightforward, thus we omit it for the sake of brevity.

The connections in a SBNN are the mapped one-valued weights, i.e. the set $S_1$. Therefore, SBNNs do not require any $\mathrm{xnor}$ operation, being $\mathrm{popcount}$ the only bitwise operation needed during the forward pass. The latter, however, is performed only in a layer's input bits connected through the one-valued weights rather than the full input. 

For any given layer $\ell$, the number of binary operations of a BNN in such layer is $\mathcal{O}_{\text{\tiny{BNN}}} = 2 N_\ell$ \cite{bethge2019backtosimplicity}. A rough estimate of the implementation gain in terms of the number of binary operations of SBNNs w.r.t. BNNs can be expressed in terms of the EC as
\begin{equation}
  \dfrac{\mathcal{O}_{\text{\tiny{SBNN}}}}{\mathcal{O}_{\text{\tiny{BNN}}}} \approx \dfrac{2N_\ell}{\text{EC}\cdot N_\ell} \approx \dfrac{2}{\text{EC}}. 
\end{equation}

This term indicates that the lower the EC fraction, the higher the gain w.r.t. BNNs. \Cref{implementation_fig} illustrates this gain. 

To illustrate the gains in the SBNN's operations at inference time, let us generalize $\mathbf{x}$ to be the input vector to any layer and $\mathbf{z}=\mathbf{w}\,\mathbf{x}$ its output. Using \Cref{alphabeta_to_zeroone_eq}, $\mathbf{z}$ can be computed as
\begin{equation} \label{implementation_eq}
\mathbf{z} = \beta'\,\mathbf{z}' + \beta'\,\alpha'\,\mathbf{q},
\end{equation}
where $\mathbf{z}'=\mathbf{w'}\,\mathbf{x}$ and $\mathbf{q} = \mathbf{1}\,\mathbf{x}$, with $\mathbf{1}$ all-ones matrix. The product to obtain $\mathbf{z}'$ requires only $\mathrm{popcount}$ operations and no $\mathrm{xnor}$ operation (\Cref{implementation_fig}), whereas all the elements in $\mathbf{q}$ take the value $2\cdot\mathrm{popcount}(\mathbf{x}) - |\mathbf{x}|$, with $|\mathbf{x}|$ the size of $\mathbf{x}$. Therefore, they are computed only once. Finally, $\mathrm{batchnorm}$ and $\mathrm{sign}$ operations following the estimation of $\mathbf{z}$ can be efficiently implemented as a threshold comparison \citep{umuroglu2017finn}, as detailed in \Cref{implementation_appendix_sec}.

\subsection{Design Tools}\label{design_sec}
In SBNN design, it is important to consider the resource constraints of the hardware where these networks are to be deployed~\citep{lin2020mcunet}. For instance, given a model topology, it is important to establish what is the compression rate it can achieve and if it meets the storage capacity of the hardware. 
The compression rate (CR) is defined as the full-precision to compressed SBNN model size rate, i.e.
\begin{equation}\label{eq:comp_rate}
  \text{CR}_{\text{\tiny{A}}} = \frac{\mathcal{W}_{\text{\tiny{FP}}} + \overline{\mathcal{W}_{\text{\tiny{FP}}}}}{\mathcal{W}_{\tiny{\text{\tiny{SBNN,A}}}} + \overline{\mathcal{W}_{\text{\tiny{FP}}}}},
\end{equation}
where $\mathcal{W}_{\tiny{\text{\tiny{SBNN,A}}}}$ is the size in bits of the compressed SBNN model using encoding algorithm $A$, whereas $\mathcal{W}_{\text{\text{\tiny{FP}}}}$ is the equivalent full precision model. Since not all layers of the full precision DNN are compressed (e.g. batch normalization layers), $\overline{\mathcal{W}_{\text{\tiny{FP}}}}$ denotes the size in bits of the uncompressed full-precision layers. 

While both $\mathcal{W}_{\text{\text{\tiny{FP}}}}$ and $\overline{\mathcal{W}_{\text{\tiny{FP}}}}$ are straightforward to compute, estimating a CR bound for a compression algorithm accounts to finding an expression for $\mathcal{W}_{\tiny{\text{\tiny{SBNN,A}}}}$. We formulate compression rate bounds of SBNN using two encoders commonly used for data 
compression: the index encoder and the run-length encoder. The IE encodes the indexes of the ones values in the SBNN, whereas RLE encodes the run-length of zeros of the binary matrices of the SBNN. Both algorithms are detailed in \Cref{compression_sec}.



\begin{assumption}\label{ass:bernoulli}
The binary weights of a SBNN follow a Bernoulli distribution of parameter $p$, i.e. 
\begin{equation*}
  P(w'_{i} = 1) \sim \mathcal{B}(p), 
\end{equation*}
with $p=$EC.
\end{assumption}
\begin{assumption}\label{ass:smooth}
For every layer $\ell$ in a SBNN, the fraction of non-zero weights is equal to the network's expected connections (EC), i.e. $ |\smash{S_{1^{(\ell)}}}|/N_{\ell} = EC, \,\, \forall \, \ell$.
\end{assumption}
\begin{proposition}\label{def:ie}
The expected size in bits of encoded SBNN's weights using index encoding (IE), $\mathcal{W}_{\text{\tiny{SBNN,IE}}}$, is expressed as a function of the EC:
\begin{equation}\label{index_encoder_size_eq}
\mathcal{W}_{\text{\tiny{SBNN,IE}}} = \sum_{\ell=1}^{L^*} n^{(\ell)}_{\text{bits,IE}} \cdot \text{EC} \cdot N_{\ell} + (n^{(\ell)}_{\text{bits,IE}} + 1) \cdot \frac{N_{\ell}}{n^{(\ell)}_d} + 16 \cdot D_{\ell} + 32 \cdot 2
\end{equation}
with $\ell$ an index of the $L^*$ SBNN's compressed layers, 
\begin{equation*}
n^{(\ell)}_{\text{bits,IE}} = \left\lceil\log_2(n^{(\ell)}_d)\right\rceil
\end{equation*}
the number of bits used to encode every single 1 index, 
$n^{(\ell)}_d$ the $d^{th}$ dimension size among the $D_{\ell}$ dimensions of the weight matrix of the $\ell^{th}$ network layer, $16 \cdot D_{\ell}$ is the number of bits used to store the length of each dimension, and $32 \cdot 2$ are the bits used to store the learned parameters $\alpha'^{(\ell)}$ and $\beta'^{(\ell)}$.
\end{proposition}
\textbf{Proof:} See \Cref{IE_proof_sec} $\qedsymbol$
\begin{proposition}\label{def:rle}
The minimum size in bits of the compressed SBNN weight matrices using RLE, $\mathcal{W}_{\text{\tiny{SBNN,RLE}}}$, is given by
\begin{equation}\label{runlength_encoder_size_eq}
\mathcal{W}_{\text{\tiny{SBNN,RLE}}} = \sum_{\ell}^{L^*} n^{(\ell)}_{\text{\tiny{bits,RLE}}} \cdot \lfloor\text{EC} \cdot N_{\ell}\rfloor + 32 \cdot n^{(\ell)}_1 + 16 \cdot (D_{\ell}+1) + 32 \cdot 2
\end{equation}
with
\begin{equation*}
n^{(\ell)}_{\text{\tiny{bits,RLE}}} = \left\lceil\log_2 \left\lceil\frac{N_{\ell} - \lfloor\text{EC} \cdot N_{\ell}\rfloor}{\lfloor\text{EC} \cdot N_{\ell}\rfloor - 1}\right\rceil\right\rceil
\end{equation*}
the number in bits of every run-length of zeros, $n^{(\ell)}_d$ the number of sub-matrices encoded via the RLE, and $16 \cdot (D_{\ell}+1)$ the length in bits used to store the length of each dimension and to represent every run-length. 
\end{proposition}

\textbf{Proof:} See \Cref{RLE_proof_sec} $\qedsymbol$
	\section{Experiments and Results}\label{sec:results}
	\paragraph{Datasets.} We evaluated the proposed SBNN on three classification benchmarks widely used to study BNNs \cite{Courbariaux2015,Courbariaux2016,courbariaux2017,bitwise, ding2019regularizing}: \textbf{1) the MNIST dataset} \cite{MNIST}. It comprises of 60K 28$\times$28 black and white images of handwritten digits with labels for 10 classes; \textbf{2) CIFAR-10} and \textbf{3) CIFAR-100} \cite{Cifar10}, consisting of 50K 32$\times$32 RGB images of natural scenes with 10 and 100 classes, respectively. All datasets have 10K images for testing. 
	
	\paragraph{Setup.} We used the BNN training algorithm proposed by \citet{Courbariaux2016} as the backbone for our SBNN training scheme. All experiments have used $\textrm{Adamax}$ \citep{kingma2014adam} as optimizer and we trained each experiment using 1 GPU from the Google Colaboratory service.
	
	We compare the performance of our SBNN w.r.t. (non-sparse) BNN formulations in terms of accuracy and compression performance. The compression performance of SBNNs and BNNs is measured in terms of the compression rate. We use three different encoders to compress the network: the index encoder (IE), run-length encoder (RLE) and huffman encoder (HE); and we compare them with the case with no encoder (NE), where every binary parameter is represented with 1 bit. We chose to include HE, despite not providing compression bounds for it, as it is commonly used in the literature. Details on the compression procedure are found in \Cref{compression_sec}.
	
	
\subsection{SBNN Performance}
We study the performance of SBNNs, in terms of accuracy and CR, when using different backbone network architectures and different EC values. To this end, we study SBNNs using linear layers, using the MNIST dataset and SBNNs with convolutional layers over CIFAR-10 and CIFAR-100. 

\begin{figure}[t]
  \centering
  \subfloat[Linear]{\includegraphics[width=0.28\textwidth]{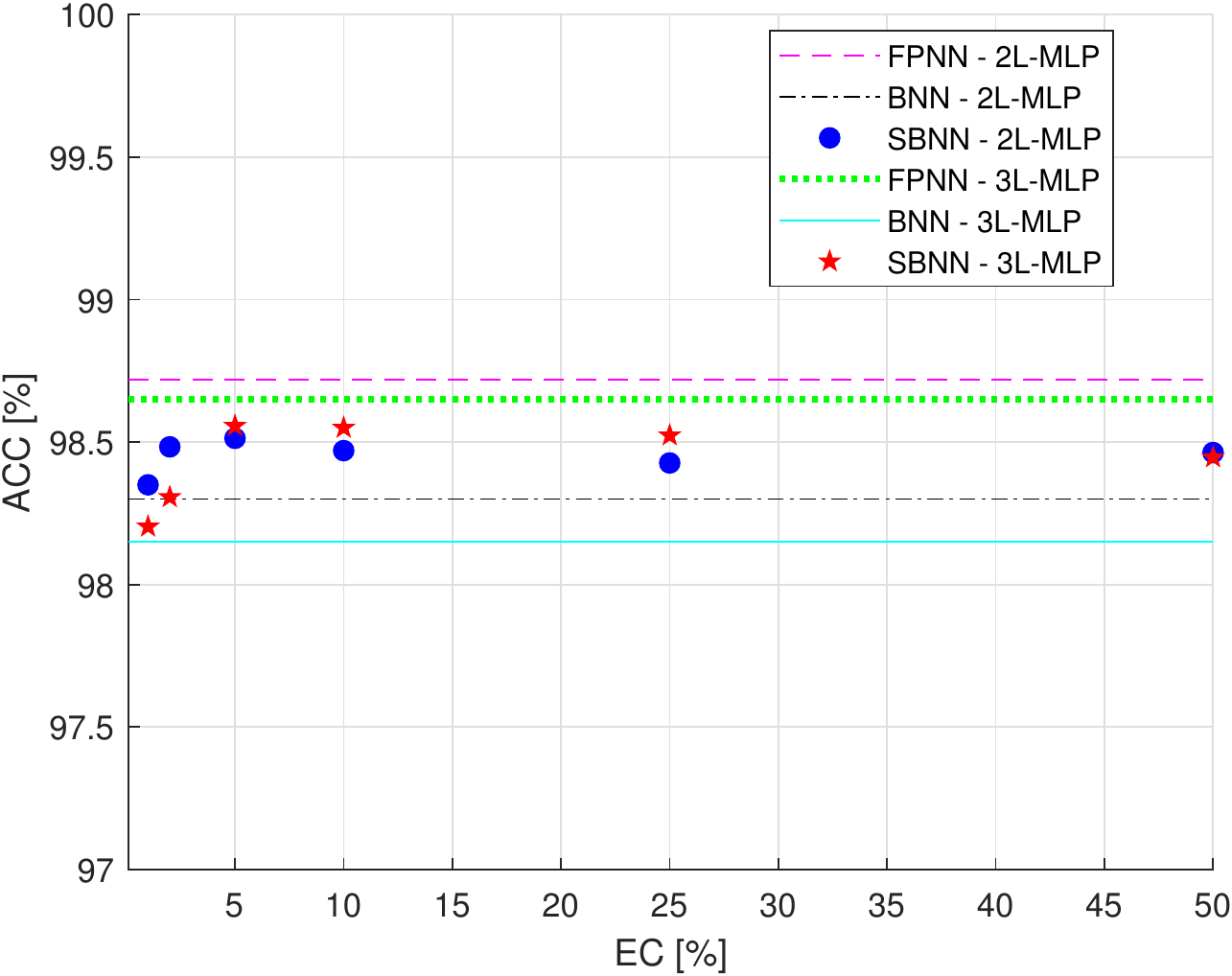}} \,\, \qquad
		 \subfloat[Convolutional ]{\includegraphics[width=0.28\textwidth]{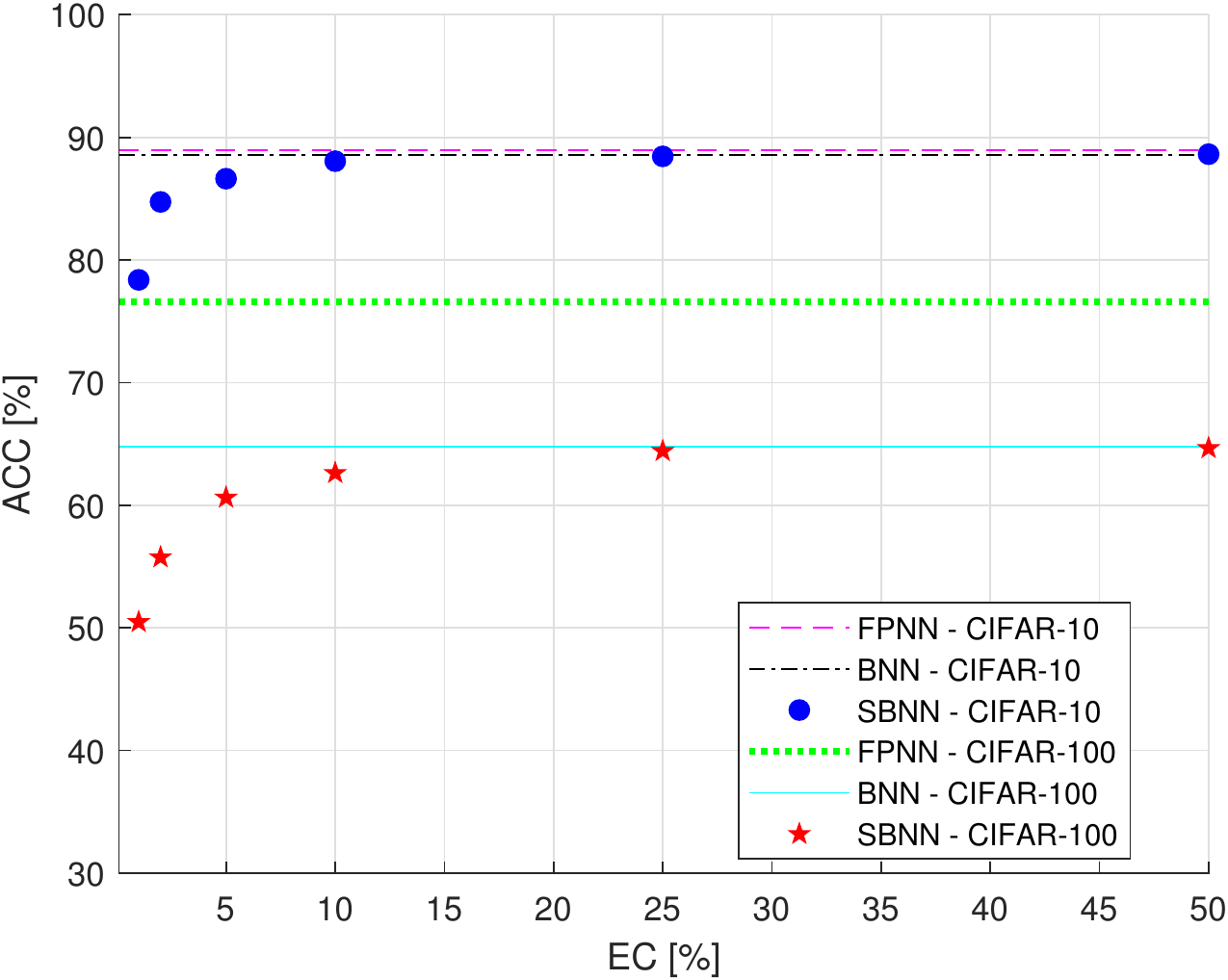}}
  \caption{Accuracy vs. EC for a SBNN with (a) linear and (b) convolutional topologies, and the equivalent full-precision neural network (FPNN) and binary neural network (BNN).}
 		\label{linear_simulation}
\end{figure}

\textbf{Linear Layers.} We use two linear topologies, consisting of 2 and 3 hidden layers with 1024 neurons each and we denote these architectures as 2L-MLP and 3L-MLP, respectively. Further implementation details are provided in \Cref{app:linear}.

\Cref{MNIST_results_table} summarizes accuracy, CR and average final model size of the best compressing algorithm (in Bytes) for both SBNNs, when varying the EC, and compares them with the full-precision equivalent network, as well as, BNN \citep{Courbariaux2016} trained under same conditions and topologies.

The results show that the accuracy of the different SBNN architectures, i.e. their generalization capacity, is comparable to that one of BNNs, while significantly increasing the CR and, thus decreasing the final model size. With only $1\%$ of connections, SBNNs can achieve nearly same accuracy performance, being approximately $270$ times smaller than the equivalent full-precision model and more than $8$ times smaller than the equivalent BNN model. This result is further illustrated in \Cref{linear_simulation}(a), where the achieved accuracies by SBNN with 2L-MLP and 3L-MLP over different ECs are always superior to their BNN counterpart, meaning their performance degradation w.r.t full-precision networks (FPNN) is milder.


	
	\begin{table}[t]
		\caption{Method comparison on MNIST dataset. Values in parentheses report standard deviation over 3 runs.
		}
		\label{MNIST_results_table}
		\centering
		  \resizebox{\linewidth}{!}{%
			\begin{tabular}{lr|rrrrrr|rrrrrr}
			\hline
			\multicolumn{1}{c}{ } & \multicolumn{1}{c|}{ } & \multicolumn{6}{c|}{\textbf{MNIST}, 2L-MLP} & \multicolumn{6}{c}{\textbf{MNIST}, 3L-MLP}\\
				\multicolumn{1}{c}{Model} & \multicolumn{1}{c|}{EC [$\%$]} & \multicolumn{1}{c}{ACC [$\%$]} & \multicolumn{4}{c}{Compression rate} & \multicolumn{1}{c|}{size} & \multicolumn{1}{c}{ACC [$\%$]} & \multicolumn{4}{c}{Compression rate} & \multicolumn{1}{c}{size} \\
				\cline{4-7}\cline{10-13}
				& & & NE & IE & RLE & HE & & & NE & IE & RLE & HE & \\
				\hline
				\hline
			
				{SBNN \textbf{[our]}} & $1$ & $98.35(\pm 0.04)$ & 32 & 216 & 219 & 267 & 27 kB & $98.20(\pm 0.02)$ & 32 & 219 & 220 & 275 & 41 kB\\
				\hline
				{SBNN \textbf{[our]}} & $2$ & $98.48(\pm0.15)$ & 32 & 129 & 151 & 184 & 40 kB & $98.31(\pm 0.13)$ & 32 & 130 & 152 & 186 & 61 kB\\
				\hline
				{SBNN \textbf{[our]}} & $5$ & $98.51(\pm0.08)$ & 32 & 59 & 87 & 104 & 70 kB & $98.56(\pm 0.07)$ & 32 & 59 & 87 & 104 & 109 kB \\
				\hline
				{SBNN \textbf{[our]}} & $10$ & $98.47(\pm0.13)$ & 32 & 31 & 57 & 67 & 108 kB & $98.55(\pm 0.08)$ & 32 & 31 & 56 & 66 & 173 kB \\
				\hline
				{SBNN \textbf{[our]}} & $25$ & $98.43(\pm0.04)$ & 32 & 13 & 33 & 40 & 183 kB & $98.52(\pm 0.02)$ & 32 & 13 & 32 & 38 & 298 kB\\
				\hline
				{SBNN \textbf{[our]}} & $50$ & $98.46(\pm0.11)$ & 32 & 6 & 23 & 31 & 228 kB & $98.45(\pm 0.04)$ & 32 & 6 & 23 & 31 & 356 kB \\
				\hline
				{BNN} & $100$ & $98.30(\pm0.08)$ & 32 & - & - & - & 228 kB & $98.15(\pm 0.07)$ & 32 & - & - & - & 356 kB \\
				\hline
				Full-precision & $100$ & $98.72(\pm 0.07)$ & 1 & - & - & - & 7.1 MB & $98.65(\pm 0.04)$ & 1 & - & - & - & 11.1 MB \\
				\hline
				
			\end{tabular}
			}
			
	\end{table}

\textbf{Convolutional Layers.} For CIFAR-10 we use a convolutional topology inspired on the VGG network \cite{Simonyan2014}, while for CIFAR-100 we use the pre-activation residual network \cite{he2016identity}. During CIFAR-100 training we use the $\textrm{mixup}$ strategy \citep{zhang2017mixup} to reduce overfitting. Both networks are trained for 300 epochs and a more detailed description of the network architectures and the training procedures are provided in \ref{app:convolutional}.
  
		
		\Cref{CIFAR_results_table} reports accuracy, CR and final model size (in Bytes) for SBNNs with varying EC, its full-precision equivalent network, and the BNN proposed in \cite{Courbariaux2016} when trained using CIFAR-10 and CIFAR-100. We observe that, in more complex architectures, the drop in accuracy of SBNNs with respect to BNNs significantly increases as the SBNN becomes more sparse (lower EC). Nevertheless, on CIFAR-10 it is possible to design SBNNs with $5\%$ and $10\%$ of connections which show an accuracy drop of only $1.92\%$ and $0.50\%$ w.r.t. BNNs, while reporting a compression of the full-precision network of up to 115 and 72 times, respectively. On CIFAR-100, we can compress the model up to 43 and 71 times, by designing an SBNN with $25\%$ and $10\%$ EC, while incurring in accuracy losses of $0.36\%$ and $2.16\%$, respectively.
		

	\begin{table}[t]
		\caption{Method comparison on CIFAR-10 and CIFAR-100 datasets. Values in parentheses report standard deviation over 3 runs.
		}
		\label{CIFAR_results_table}
		\begin{center}
		  \resizebox{\linewidth}{!}{%
			\begin{tabular}{lr|rrrrrr|rrrrrr}
			\hline
			\multicolumn{1}{c}{ } & \multicolumn{1}{c|}{ } & \multicolumn{6}{c|}{\textbf{CIFAR-10}, VGG} & \multicolumn{6}{c}{\textbf{CIFAR-100}, PreActResNet}\\
				\multicolumn{1}{c}{Model} & \multicolumn{1}{c|}{EC [$\%$]} & \multicolumn{1}{c}{ACC [$\%$]} & \multicolumn{4}{c}{Compression rate} & \multicolumn{1}{c|}{size} & \multicolumn{1}{c}{ACC [$\%$]} & \multicolumn{4}{c}{Compression rate} & \multicolumn{1}{c}{size} \\
				\cline{4-7}\cline{10-13}
				& & & NE & IE & RLE & HE & & & NE & IE & RLE & HE & \\
				\hline
				\hline
			
				{SBNN \textbf{[our]}} & $1$ & $78.38(\pm 0.31)$ & 32 & 260 & 321 & 347 & 52 kB & $50.47(\pm 0.62)$ & 32 & 235 & 268 & 303 & 145 kB\\
				\hline
				{SBNN \textbf{[our]}} & $2$ & $84.73(\pm0.82)$ & 32 & 157 & 194 & 214 & 84 kB & $55.75(\pm 0.63)$ & 32 & 147 & 175 & 199 & 220 kB\\
				\hline
				{SBNN \textbf{[our]}} & $5$ & $86.63(\pm1.05)$ & 32 & 71 & 103 & 115 & 155 kB & $60.62(\pm 0.62)$ & 32 & 69 & 96 & 110 & 400 kB \\
				\hline
				{SBNN \textbf{[our]}} & $10$ & $88.05(\pm0.09)$ & 32 & 37 & 63 & 72 & 249 kB & $62.62(\pm 0.73)$ & 32 & 36 & 61 & 71 & 621 kB \\
				\hline
				{SBNN \textbf{[our]}} & $25$ & $88.45(\pm0.44)$ & 32 & 15 & 34 & 41 & 435 kB & $64.42(\pm 0.50)$ & 32 & 15 & 35 & 43 & 1.01 MB\\
				\hline
				{SBNN \textbf{[our]}} & $50$ & $88.63(\pm0.20)$ & 32 & 8 & 24 & 31 & 561 kB & $64.65(\pm 0.64)$ & 32 & 7 & 24 & 32 & 1.33 MB \\
				\hline
				{BNN} & $100$ & $88.55(\pm0.27)$ & 32 & - & - & - & 561 kB & $64.78(\pm 0.24)$ & 32 & - & - & - & 1.34 MB \\
				\hline
				Full-precision & $100$ & $88.94(\pm 0.77)$ & 1 & - & - & - & 17.5 MB & $76.60(\pm 0.27)$ & 1 & - & - & - & 43 MB \\
				\hline
				
			\end{tabular}
			}
			
		\end{center}
	\end{table}
		
	
	

	\textbf{Role of $\mathbf{\gamma}$}.
	The hyperparameter $\gamma$ (\Cref{gamma_eq}) plays an important role in the SBNN training. It controls the contribution of the penalization term $\lambda\, h(\mathbf{w}')$, which controls the sparsity. To provide further insights on its role and how its influence on the achievement of a desired sparsity level, \Cref{CIFAR_gamma_table} reports a range of $\gamma$ values and the corresponding achieved number of connections for the different SBNNs. As one should expect, higher $\gamma$ values lead to higher sparsity but, as previously shown, it comes at the cost of a drop in accuracy.
	
	
	\begin{table}[t]
		\caption{Value of the hyperparameter $\gamma$ set in the experiments to achieve the desired value of EC
		}
		\label{CIFAR_gamma_table}
		\begin{center}
		  \resizebox{0.9\linewidth}{!}{%
			\begin{tabular}{lccccccc||lccccccc}
			\hline
			    & &\multicolumn{6}{c}{EC [$\%$]} & & &\multicolumn{6}{c}{EC [$\%$]}\\
			    \cmidrule(r){3-8} \cmidrule(r){11-16}
				 & \multicolumn{1}{c}{} & \multicolumn{1}{c}{1} & \multicolumn{1}{c}{2} & \multicolumn{1}{c}{5} & \multicolumn{1}{c}{10} & \multicolumn{1}{c}{25} & \multicolumn{1}{c}{50} & \multicolumn{2}{c}{ } & \multicolumn{1}{c}{1} & \multicolumn{1}{c}{2} & \multicolumn{1}{c}{5} & \multicolumn{1}{c}{10} & \multicolumn{1}{c}{25} & \multicolumn{1}{c}{50}\\
				 \hline
			   MNIST, 2L-MLP & $\gamma$ [$\%$] : & 45 & 41 & 34 & 24 & 7 & 0 & CIFAR-10 & $\gamma$ [$\%$] : & 28 & 25 & 20 & 15 & 6 & 0 \\\hline
			   MNIST, 3L-MLP & $\gamma$ [$\%$] : & 46 & 43 & 34 & 26 & 8 & 0& CIFAR-100 & $\gamma$ [$\%$] : & 7 & 6 & 4 & 3 & 1 & 0\\
				\hline
			\end{tabular}
			}
		\end{center}
	\end{table}

		 \begin{figure}[t]
		\centering
		\subfloat[MNIST, 2L-MLP]{\includegraphics[width=0.23\textwidth]{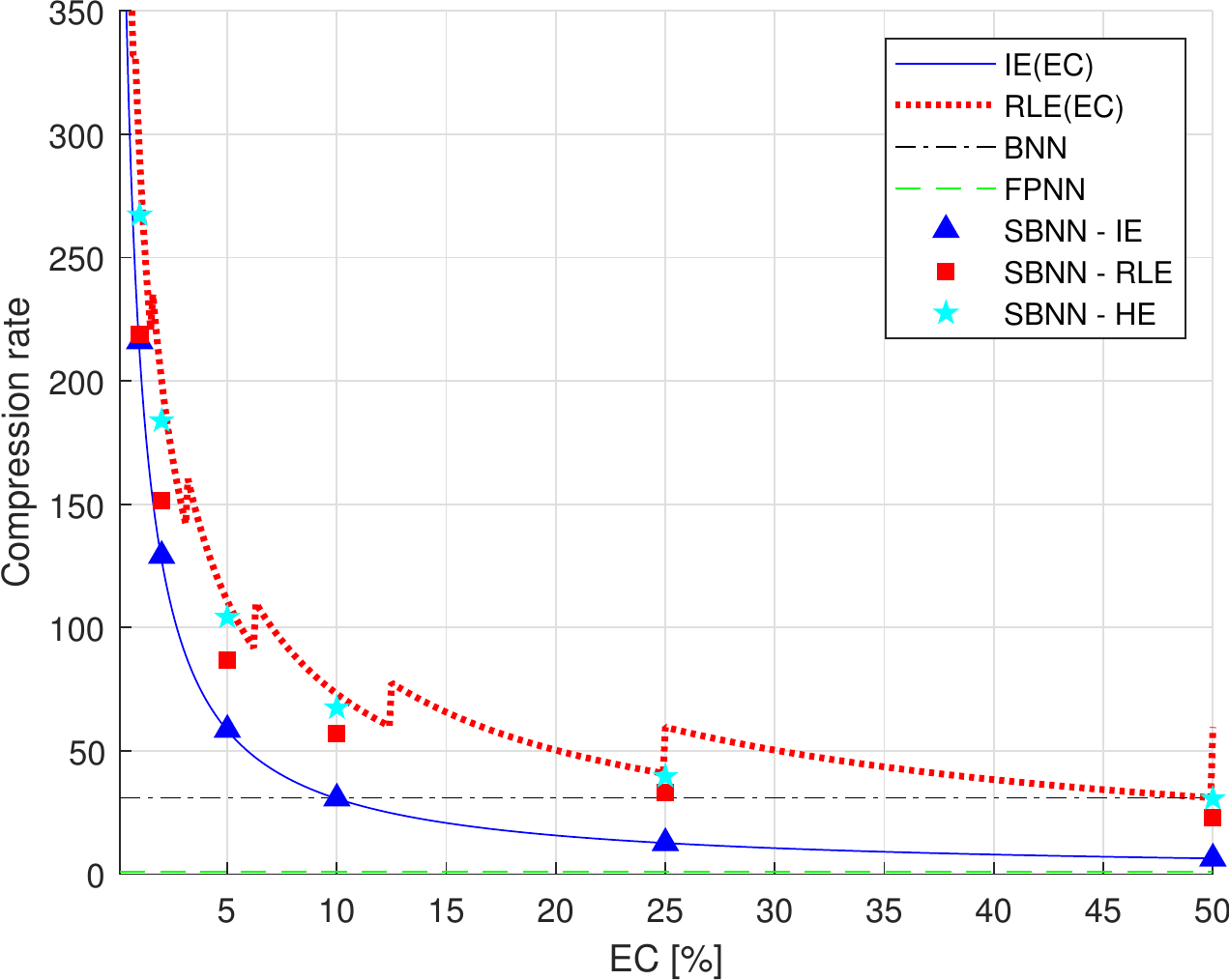}}\ 
    \subfloat[MNIST, 3L-MLP]{\includegraphics[width=0.23\textwidth]{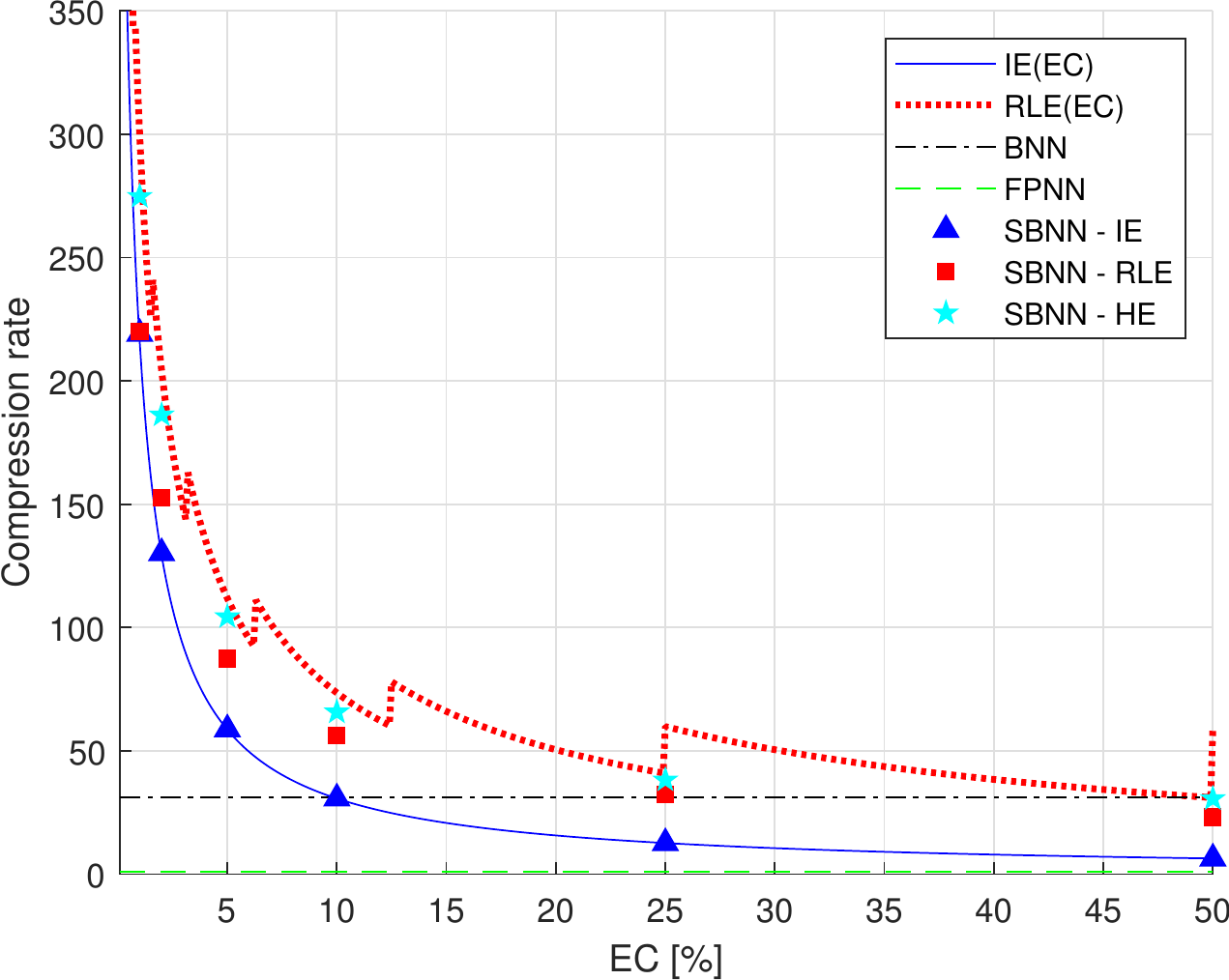}}\
		 \subfloat[CIFAR-10]{\includegraphics[width=0.23\textwidth]{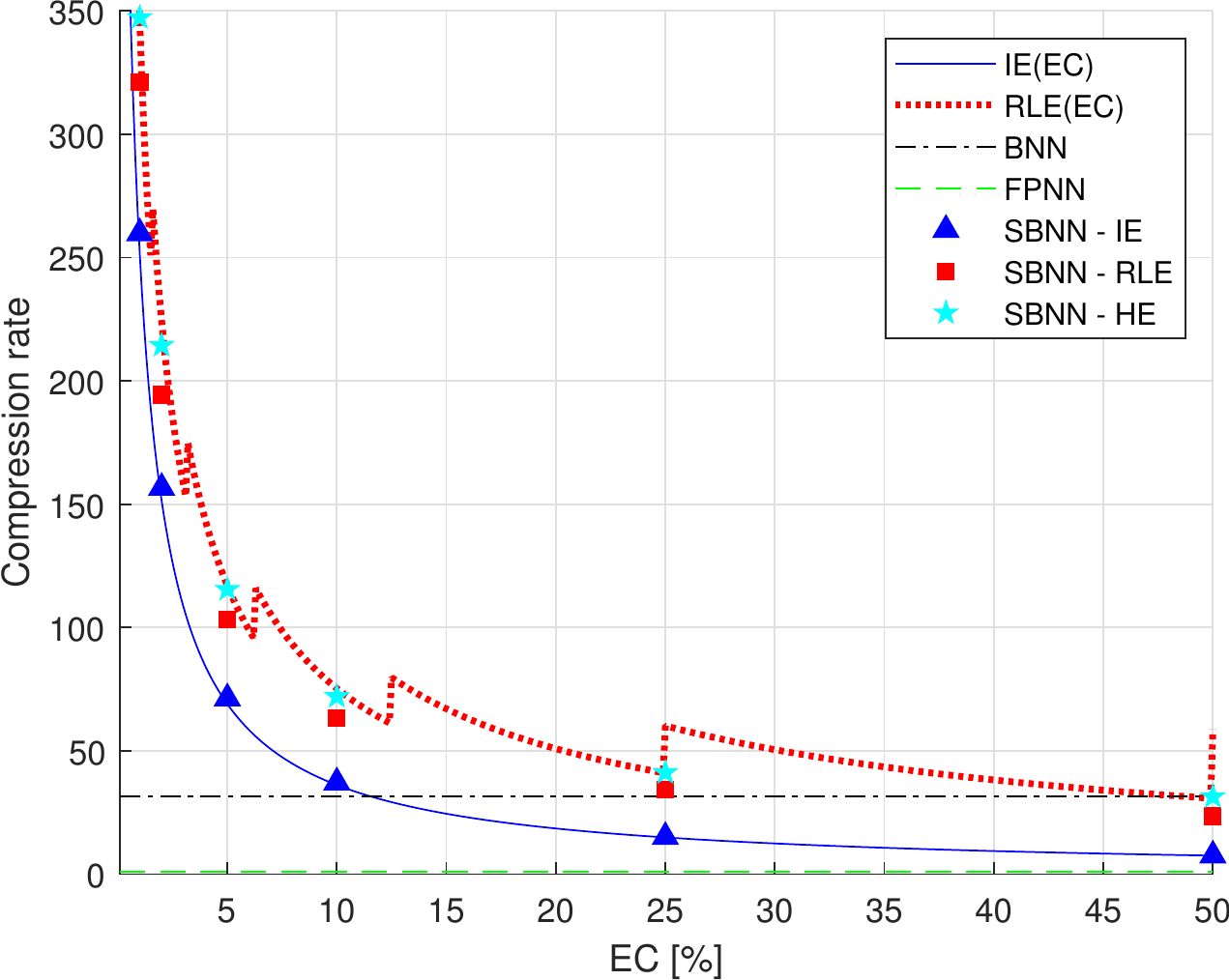}}\ 
		 \subfloat[CIFAR-100]{\includegraphics[width=0.23\textwidth]{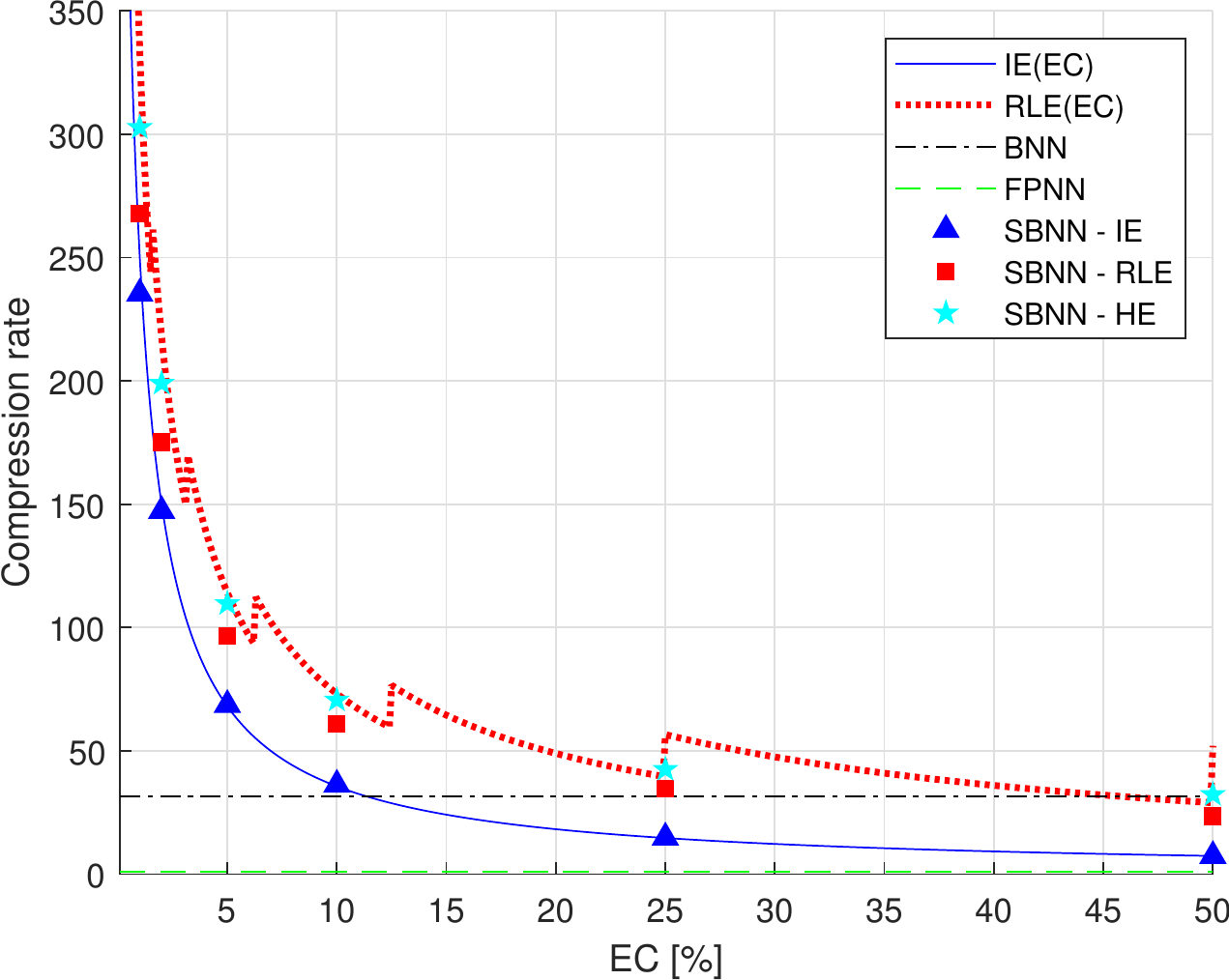}}\

 		\caption{Achieved compression rate vs. EC for a SBNN with a linear topology trained on MNIST (a) 2L-MLP and (b) 3L-MLP; a convolutional topology trained on (c) CIFAR-10 and (d) CIFAR-100, the respective equivalent full-precision neural networks (FPNN) and binary neural networks (BNN).}
 		\label{conv_simulation}
 	\end{figure}	
 		\begin{figure}[!t]
		\centering
		 \subfloat[MNIST, 2L-MLP]{\includegraphics[width=0.23\textwidth]{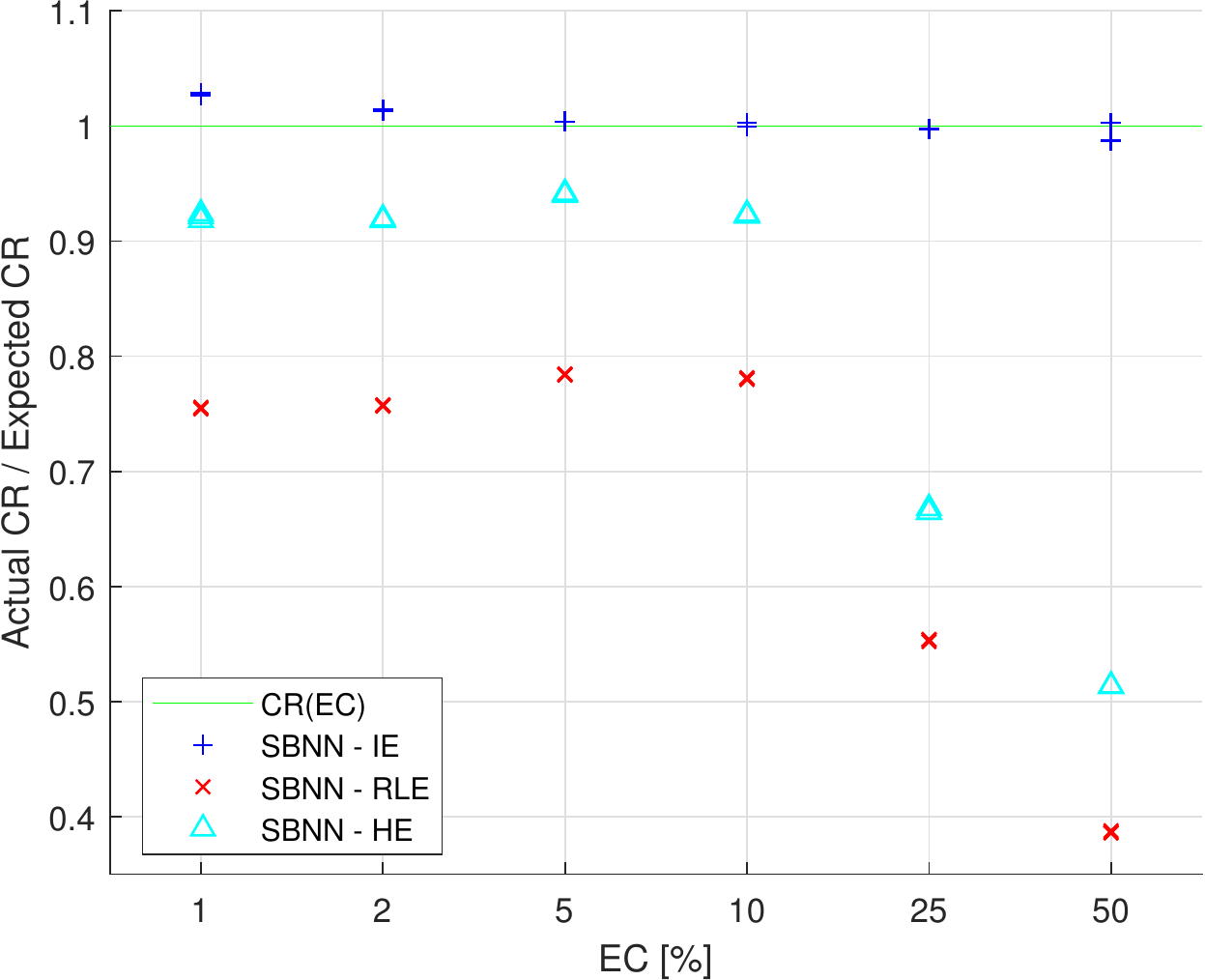}}\ \ 
		 \subfloat[MNIST, 3L-MLP]{\includegraphics[width=0.23\textwidth]{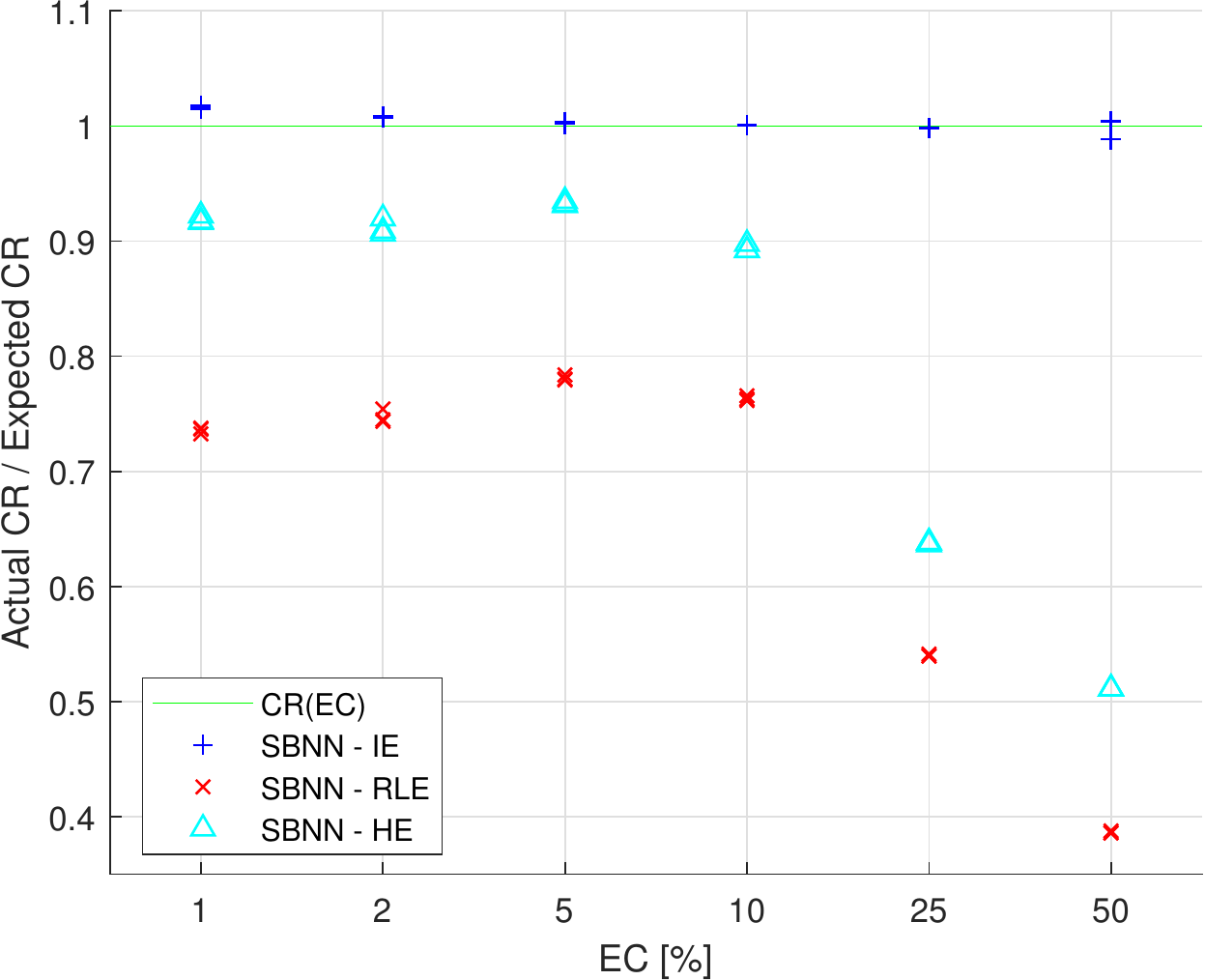}}\ \ 
		 \subfloat[CIFAR-10]{\includegraphics[width=0.23\textwidth]{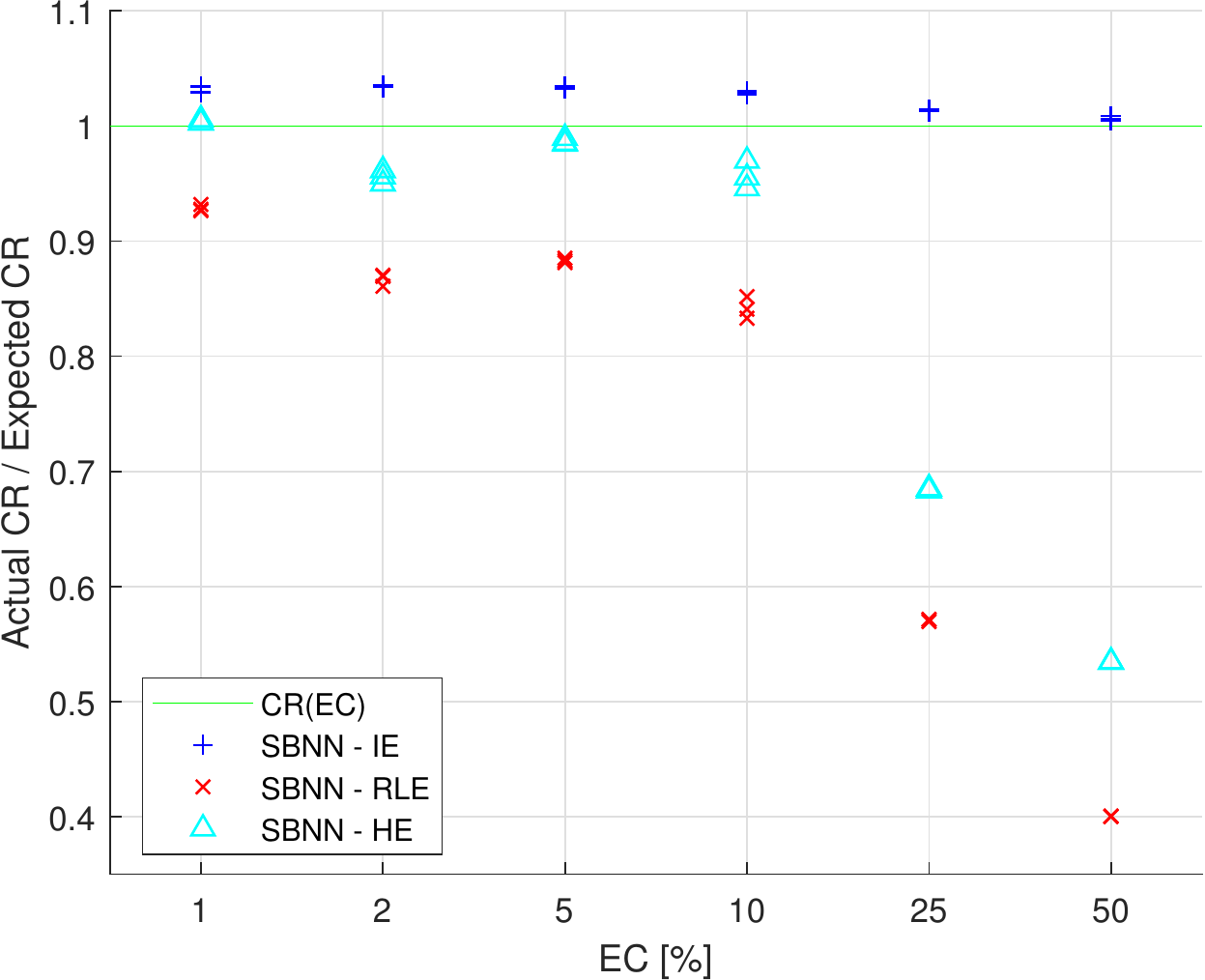}}\ \ 
		 \subfloat[CIFAR-100]{\includegraphics[width=0.23\textwidth]{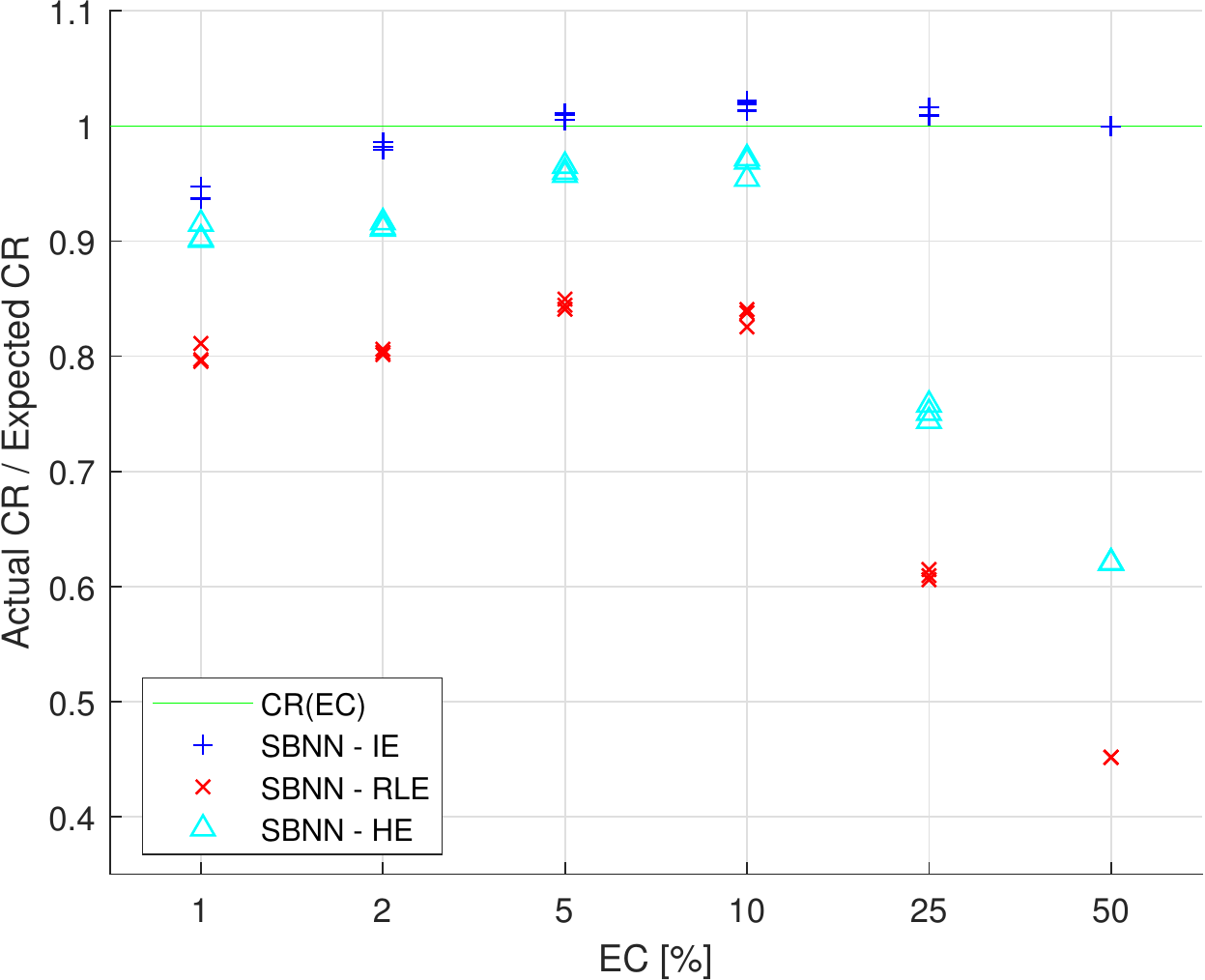}}
 		\caption{Achieved CR to estimated CR bound ratio for different ECs over 3 different runs on MNIST (a) with 2L-MLP and (b) 3L-MLP, on (c) CIFAR-10 and on (d) CIFAR-100 with IE ({\color{blue}$+$}), RLE ({\color{red}$\times$}) and HE ({\color{cyan}$\triangle$}).}
 		\label{design_precision_fig}
 	\end{figure}

	\subsection{Design Tools Validation}
	We validate the proposed CR bounds in \Cref{conv_simulation} by comparing the estimated theoretical values for the four SBNN topologies so far studied and by comparing these to the measured compression rate after SBNN training. \Cref{conv_simulation} plots the CR bound curves (IE and RLE) as EC varies along with the measures CR after network 
	encoding, showing a good correlation between estimated and measured values. 
	
	To further study the accuracy and reliability of the formulated CR bounds, we measured the ratio between the measured CR in an experiment and the expected CR bound IE and the upper CR bound RLE (\Cref{design_sec}). The closer the ratio is to $1$, the higher the accuracy and reliability of the CR bounds as SBNN design tools. Moreover, we study also the ratio of the measured CR of the HE and the upper CR bound of the RLE, to see if such bound can be used to predict also the HE compression. \Cref{design_precision_fig} reports the results obtained on the fours studied topologies for different EC values. The ratio achieved with IE is always very close to 1 with deviations from it below $5\%$. For EC values smaller or equal than $10\%$, the ratios achieved with RLE get close to 1. However, compared to IE, we observe larger deviations from the upper bound: up to $\sim$30\% in linear topologies and up to $\sim$20\% in convolutional ones. It is also interesting to note that the CR upper bound built on the RLE is a good estimate of the real compression bound when the HE is used, and for values of EC smaller or equal than $10\%$, the deviation of HE from the bound is up to $\sim10\%$.

	\section{Conclusions}\label{sec:conclusions}
	
	We proposed sparse binary neural network (SBNN), a method to design and further compress BNNs by introducing sparsity and reducing the required computations. Our approach is based on the quantization of weights in a general $\alpha$/$\beta$ binary domain, which are then expressed as $0s$ and $1s$ when implemented. Our method is formulated as a mixed optimization problem and we show that it can be solved using any state-of-the-art BNN training algorithm, through the addition of two learnable parameters and a regularization term controlling sparsity. We have proposed theoretical tools to design SBNNs based on the storage capacity of the hardware where they have to be deployed. Our results over three data sets (MNIST, CIFAR-10 and CIFAR-100) and two different topologies (feed-forward linear and convolutional networks) used as backbone architectures indicate that SBNNs are a powerful and flexible method to deploy DNNs in IoT devices and sensors. For instance, a linear SBNN model with two hidden layers and a model size of 27 kB (Huffman encoder)  can be entirely stored, not only in the flash memory, but also in the SRAM memory of very low power hardware modules like the Intel\textsuperscript{\textregistered} Curie\textsuperscript{TM} Module, which has 384kB of flash memory and only 80kB of SRAM, while achieving $\sim 98.35 \%$ accuracy on MNIST. A ResNet-like SBNN topology with $\sim 64.42 \%$ accuracy on CIFAR-100 can be stored on a STM32F746 microcontroller, which has only 1MB of storage space. The benefits of deploying DNNs on such small devices are countless: they increase data privacy, as there would be no need to send acquired data to a cloud for processing \citep{mcmahan2017communication} or decrease the response latency in autonomous systems \citep{iot_application}.

	A current weakness of SBNNs is that it requires to set an extra hyperparameter $\gamma$ w.r.t. BNNs. However, we think that further studies of $\gamma$, based on control theory approaches with dynamic adaptation of this value, may allow a lower degradation of the accuracy, while extending the compression capabilities of current SBNNs. Future works should focus on this topic and on possible frameworks to provide fast and open hardware implementation solutions for SBNNs.

\bibliography{neurips2021}

\input{neurips2021_appendix}
\end{document}

%% file: neurips2021_appendix.tex
\appendix

\section{Appendix}

	\subsection{SBNN Implementation: Batchnorm layer an activation function}\label{implementation_appendix_sec}
	We provide details about the implementation of the $\mathrm{batchnorm}$ layer and the $\mathrm{sign}$ activation function that follow the estimation of $\mathbf{z}$ in a SBNN (\Cref{implementation_sec}). Implemented in a similar fashion as BNNs, the operations of the $\mathrm{batchnorm}$ layer and the $\mathrm{sign}$ function can be merged together and implemented through an element-wise threshold comparison of the output of the linear or convolutional layers \citep{umuroglu2017finn}. This strategy avoids operations like multiplications, additions, and mean and standard deviation computations. 
    Let us denote $o_i$ the $i^{th}$ output of the $\mathrm{sign}$ activation function, i.e.,
	
	\begin{equation}\label{batchnorm_implementation_others_eq}
 o_i = \left\{\begin{array}{ll}
  +1, & \text{if } z_i \gtreqqless \delta_{i}\\
  -1, & \text{otherwise}
 \end{array}\right.\,,
 \end{equation}
    where $z_i$ is the $i^{th}$ element of $\mathrm{z}$ (\cref{implementation_eq}), while $\delta_{i}$ 
    is the threshold of the corresponding comparator. {This threshold does not correspond to the zero-value from the $\mathrm{sign}$ function. Instead, it is moved according to the values learned by the $\mathrm{batchnorm}$ parameters which follows $z_i$ \citep{umuroglu2017finn}. Therefore, } the comparison operator ($\geq$ or $\leq$) depends on the sign of the $\mathrm{batchnorm}$ parameters. Both the threshold $\delta_{i}$ and the comparison operator are computed after training and their values stored.
    
    Using \Cref{implementation_eq}, we can rewrite $o_i$ as
    
    \begin{equation}\label{batchnorm_implementation_eq}
 o_i = \left\{\begin{array}{ll}
  +1, & \text{if } z'_i \gtreqqless \frac{\delta_{i}}{\beta'} - \alpha'\,q_i\\
  -1, & \text{otherwise}
 \end{array}\right..
 \end{equation}
 
To save computations, the term $\delta_{i}/\beta'$ is computed after training and stored
, whereas $\alpha'\,q_i$ is the same value for any $i$ (see \Cref{implementation_sec}). In practice, this means that the operations involving $\alpha'$ and $\beta'$ parameters, and the $\mathrm{batchnorm}$ and $\mathrm{sign}$ operations are implemented with one addition and a comparison.
	

	
	\subsection{Compression Techniques and CR Bound Proofs}\label{compression_sec}
	We first provide a detailed description of the the encoding techniques used in this study to compress the sparse weight matrices of the linear and convolutional layers of the SBNNs. These are: index encoder (IE), run-length (RLE) and Huffman. Then, we present the proofs for the derived compresssion rate bounds (\Cref{def:ie} and \Cref{def:rle}) for IE and RLE.
	
	\paragraph{Compression techniques} A SBNN can be seen as a set of sparse binary matrices. Well-known methods from source coding theory can be exploited for the compression of sparse vectors and matrices. 
	
\begin{figure}[ht]
 		\begin{center}
 		  \includegraphics[clip, trim={0cm 13cm 0cm 0cm}, width=\textwidth]{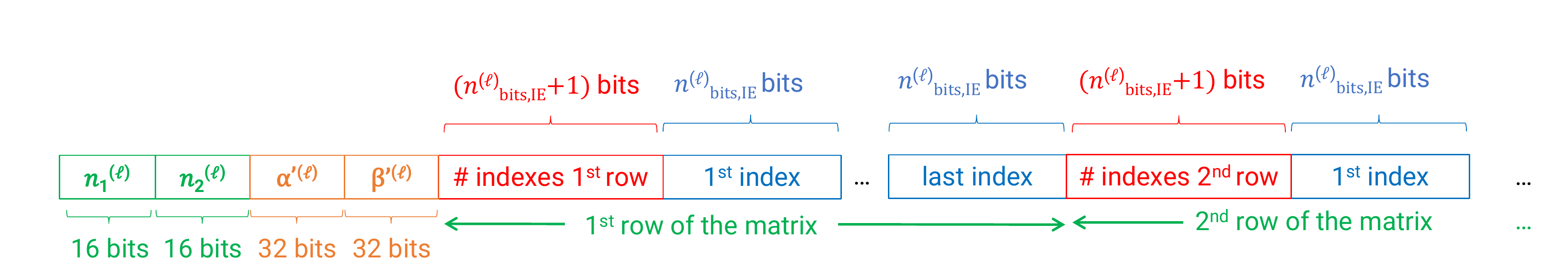}
					
 		\end{center}
 		\caption{Schematics of the IE algorithm used. The symbol $\#$ means \Quotes{number of}.}
 		\label{ie_fig}
 	\end{figure}

    The IE encodes the column (or row) indexes of the one-valued elements in a matrix. Our IE implementation, encodes the indexes of every matrix row. To delimit an encoded row's start and end we use the number of encoded indexes of every matrix row as a delimiter, even if the row has no indexes (i.e., one-valued elements). This is illustrated in \Cref{ie_fig}.
	
\begin{figure}[ht]
 		\begin{center}
 		  \includegraphics[clip, trim={0cm 10cm 0cm 0cm}, width=\textwidth]{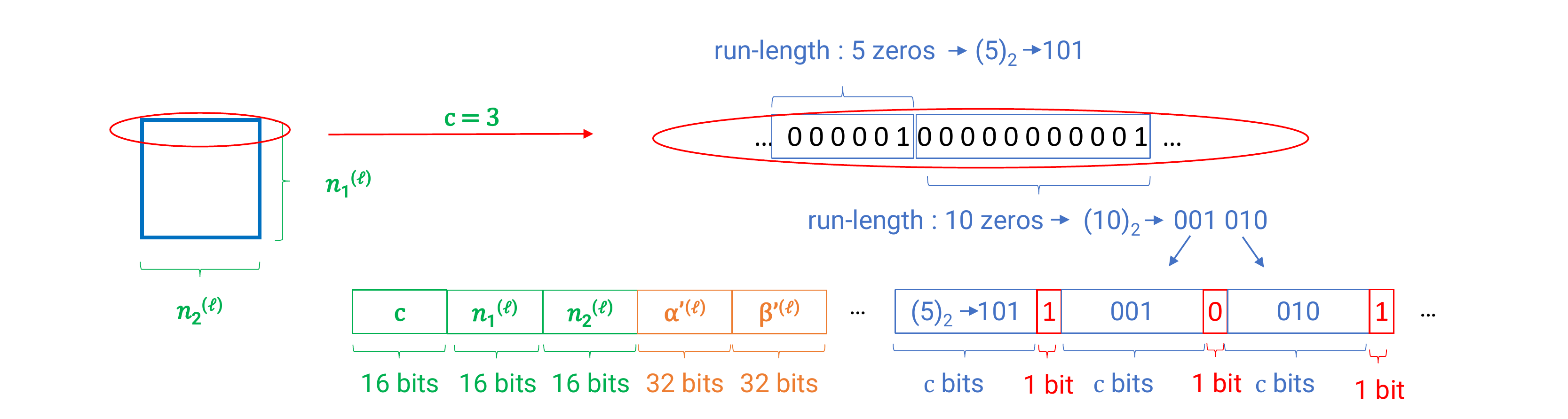}
					
 		\end{center}
 		\caption{Schematics of the RLE algorithm used.}
 		\label{rle_fig}
 	\end{figure}

	The RLE acts on runs of data, where a run denotes a sequence where the same data value occurs along several consecutive elements. Rather than storing the original run, it stores the single data value and the count, which is the number of consecutive elements also known as the run-length. The RLE is depicted in \Cref{rle_fig}. 
	
	In our work, we adapt the standard RLE algorithm to handle 0/1 values, rather than any possible value. First, we encode only the zeroes. There is no need to store the data value, given that is always the same. 
	
	Second, we set the number of bits to store the run-length count to a value $c$ that optimizes the use of the allocated $c$ bits. Since the run-length might be of varying size, it is necessary to select a number of bits for the count value which is large enough to represent the largest run-length. However, if there are only few cases with very large run-length values, many bits of the count will not be used. To avoid this phenomenon, we choose $c$ to be smaller than what it would be needed to represent the largest run-length. To encode a run-length count that can be represented with the $c$ bits, we encode it and then we append a $1$-valued bit to the $c$ bits, to indicate that the run-length count has been stored successfully. Otherwise, the run-length count is split into groups of $c$ bits. In between each $c-$bits we append a $0$-valued bit to indicate that there is at least one next set of $c$ bits encoding the count. After the last group, a $1$-valued bit is appended, thus acting as a terminator ending the count. 
	
	We choose $c$ from the set $\mathcal{C}=\{1,2,\hdots,\log_2(\text{largest run-length})\}$, after computing for every element in $\mathcal{C}$ the size in bits of the encoded sparse weight matrices of the SBNN with that value, and selecting that one leading to the smallest size in bits of the encoded SBNN.
	
	Finally, the Huffman encoder \cite{huffman} maps the elements from a set $\mathcal{A}$ to bit-symbols of variable length, using their frequency of appearance. More frequent elements are encoded using less bits, whereas less frequent elements are mapped to larger in bit symbols. In this study, we define $\mathcal{A}$ as the set of values of the run-lengths of zeros. We chose this set of elements as done in the Modified Huffman Coding standard used in fax communications \citep{recommendation1988facsimile}.
	
					
 	
 	In our experiments, we compute the full precision model size considering each weight stored with 32-bits plus 32-bits for each scaling factor introduced by the batch-normalization layers. For all encoding algorithms, we estimated the SBNN's achieved compression rates according to \Cref{eq:comp_rate}, whereas its model size was computed after compressing the binary matrices storing $\alpha'$ and $\beta'$ (\cref{alphabeta_to_zeroone_eq}) with 32 bits and storing each dimension size with 16-bits. For the RLE, we stored the run-length size with 16 bits.
 	
 	
 	\subsubsection{Index Encoder Compression Rate Bound}\label{IE_proof_sec}
 	Given Assumption~\ref{ass:bernoulli} and Assumption ~\ref{ass:smooth}, we formulate an expected bound of the CR when the IE is used. 
 	%
 	%
 	
 	Let us denote $L^* < L$ as total number of compressible linear layers in a SBNN using IE, and $\ell \in L^*$ any given compressible linear layer.  Let us also define $\smash{n_d^{(\ell)}}$ as the number of 
 	elements in the $d^{th}$ dimension of the layer's weight matrix with total number of weights $N_\ell$, i.e. 
 	\begin{equation*}
 	    N_\ell = \prod_d^{D_{\ell}} n_d^{(\ell)},
 	\end{equation*}
 	 with $D_{\ell}$ the matrix dimension. The total number of bits $n_{\text{bits,IE}}^{(\ell)}$ required to encode column indexes containing a 1 is
 	\begin{equation*}
 	    n_{\text{bits,IE}}^{(\ell)} \geq \lceil\log_2(n_d^{(\ell)})\rceil,
 	\end{equation*}
 with $\lceil \cdot \rceil$ the ceiling function. We choose $d=2$, which, in a linear topology, represents the matrix columns. Similar expressions can be found using a different value of $d$. For convenience, we constraint 
	\begin{equation}\label{eq:num_bits_ie}
 	    n_{\text{bits,IE}}^{(\ell)} = \lceil\log_2(n_2^{(\ell)})\rceil,
 	\end{equation}
	because the larger $\smash{n_{\text{bits,IE}}^{(\ell)}}$, the larger the number of bits used for encoding each index, which would result in a larger size in bits of the compressed $\ell^{th}$ layer's weigth matrix. 
	
	Given Assumption~\ref{ass:smooth} and \Cref{eq:num_bits_ie}, the total number of bits required to encode the $1-$valued weights using IE
	in $\ell^{th}$ layer is
	\begin{equation}\label{eq_ie_step1}
 	    n_{\text{bits,IE}}^{(\ell)} \cdot \text{EC} \cdot N_{\ell}.
 	\end{equation}
 	
To delimit the start and end of every encoded row, we use the number of encoded indexes of every matrix row (i.e. the number of one-valued weights) as a delimiter. Given that every row has a number of encoded indexes between 0 and $\smash{n_2^{(\ell)}}$ elements, the size in bits required to store them is 
	\begin{equation}\label{eq_ie_step2}
 	    (n_{\text{bits,IE}}^{(\ell)}+1) \cdot \frac{N_{\ell}}{n_2^{(\ell)}}. 
 	\end{equation}
	
	
	By \Cref{eq_ie_step1} and \Cref{eq_ie_step2} we obtain
	\begin{equation}\label{eq_ie_end}
 	    n_{\text{bits,IE}}^{(\ell)} \cdot \text{EC} \cdot N_{\ell} + (n_{\text{bits,IE}}^{(\ell)}+1) \cdot \frac{N_{\ell}}{n_2^{(\ell)}} + K_\ell,
 	\end{equation}
 	the number of bits for encoding in the $\ell^{th}$ layer, where
 	\begin{equation}\label{eq:overhead}
 	    K_\ell=16 \cdot D_\ell + 32 \cdot 2
 	\end{equation}
 	is a constant overhead to store the size of the $D_{\ell}$ dimensions of the weight matrix (16 bits), and the layer's $\alpha'$ and $\beta'$ parameters (32 bits each). 
 	
 	By summing over all $\ell \in L^*$, we obtain $\mathcal{W}_{\text{\tiny{SBNN,IE}}}$ (\cref{index_encoder_size_eq}), thus completing the proof.
	
	
	

		\subsubsection{Run-Length Encoder Compression Rate Bound}\label{RLE_proof_sec}
		
		We formulate an upper bound of the expected CR when the RLE is used. We use the same notation as before, with the exception of $L^*<L$ that now denotes the number of compressible linear layers in a SBNN using RLE.
	
	\begin{definition}\label{def:rl}
	A run-length length zero is a run-length.
	\end{definition}
	
	
	The number of run-lengths in $\ell^{th}$ layer is an integer $R_\ell \geq |\smash{S_{1^{(\ell)}}}|$. Given Definition~\ref{def:rl} and Assumption~\ref{ass:smooth},
	\begin{equation*}
	    R_\ell \geq | N_\ell \cdot \text{EC}|.
	\end{equation*}
	For practical convenience, we constraint $R_\ell$ to be
	%
	\begin{equation}\label{eq:r_ell}
	    R_\ell = \lfloor N_\ell \cdot \text{EC} \rfloor,
	\end{equation}
	since a larger $R_\ell$ requires a larger number of bits to encode the weight matrix. The operator $\lfloor\cdot\rfloor$ denotes the floor function. 

 	
 	
 	
 	\begin{theorem}[Generalized Pigeonhole Principle Theorem]\label{theorem:pigeonhole}
 	If $M$ objects are placed into $k$ boxes, then there is at least one box containing at least $\lceil M/k\rceil$ objects.
 	\end{theorem}
 	
 	By \Cref{eq:r_ell} and the Generalized Pigeonhole Principle Theorem (Theorem~\ref{theorem:pigeonhole}), we can express the largest length among all zero run-lengths $r_{\max}$ as 
	\begin{equation}
	    r_{\max} \geq \frac{N_\ell - R_\ell}{R_\ell}.
	\end{equation}
 	%
 	%
 	It is then straightforward to estimate the number of bits $n_{\text{bits,RLE}}^{(\ell)}$ required to encode the run-lengths of zeroes, i.e.
 	\begin{equation*}
 	    n_{\text{bits,RLE}}^{(\ell)} \geq \lceil\log_2(r_{\max})\rceil, 
 	\end{equation*}
 	and the minimum number of its required to encode the run-length of zeros using RLE in $\ell^{th}$ layer, i.e.
 	\begin{equation}\label{eq_rle_step1}
 	    n_{\text{bits,RLE}}^{(\ell)} \cdot R_\ell.
 	\end{equation}
 	
 	The number of bits for encoding in the $\ell^{th}$ layer can be expressed as
	\begin{equation}\label{eq_rle_end}
 	    n_{\text{bits,RLE}}^{(\ell)} \cdot R_\ell + 32 \cdot n_1^{(\ell)} + K_\ell,
 	\end{equation}
 	where $K_\ell$ represents the same overhead as in \Cref{eq:overhead} and the term $32\cdot n_1^{(\ell)}$ is  implementation-specific. Let us recall that in our RLE implementation we store the number of encoded run-lengths of zeros of every sub-matrix (i.e., the rows of a linear layer, the channels of a convolutional layer) to delimit the start and end of every encoded sub-matrix.
 	
 	Summing up over all $\ell \in L^*$ in \Cref{eq_rle_end} we obtain $\mathcal{W}_{\text{\tiny{SBNN,RLE}}}$ (\Cref{runlength_encoder_size_eq}), thus completing the proof.

	\subsection{Experiments Setup}
    We provide further details about network's topologies used in our experiments (\Cref{sec:results}) and their training.
    
	\subsubsection{Topologies.}\label{app:linear} 
	
	\paragraph{Linear Topologies} We use two linear architectures with 2 (2L-MLP) and 3 hidden (3L-MLP) layers, with 1024 neurons in each hidden layer, $28\times 28$ neurons in the input layer and and $10$ in the output. Every linear layer is followed by a $\mathrm{batchnorm}$ layer. For SBNNs and BNNs, the activation function is the $\mathrm{sign}$ function, except in the output layer where the $\mathrm{LogSoftMax}$ is used. For the full-precision case, $\mathrm{ReLU}$ is used in place of $\mathrm{sign}$.
	
	\paragraph{Convolutional Topologies} For CIFAR-10, we use a convolutional topology inspired on the VGG network \cite{Simonyan2014} with 6 convolutional layers and 1 linear layer, starting with 128 filters and doubling the number of filters every 2 layers. We use kernels of size $3\times 3$, while padding and stride are set equal $1$. Every convolutional layer is followed by a $\mathrm{batchnorm}$ layer and the $\mathrm{sign}$ activation function. Every linear layer is followed by a $\mathrm{batchnorm}$ layer with the $\mathrm{LogSoftMax}$ activation function. Additionally, every 2 convolutional layers we have a $\mathrm{MaxPool}$ layer with kernels of size $3\times 3$, padding of $1$ and stride of $2$. The full-precision network uses $\mathrm{ReLU}$ in place of $\mathrm{sign}$. 
	
	For CIFAR-100, we use the same convolutional topology as in \citep{ding2019regularizing}, which is inspired on the pre-activation residual network \cite{he2016identity}. Differently from \citep{ding2019regularizing}, we use depth 18, initial width of $64$ and final width of $512$. The full-precision network uses $\mathrm{ReLU}$ in place of $\mathrm{sign}$.

	\subsubsection{Setup.}\label{app:convolutional} 
	We use the BNN training algorithm proposed by \citet{Courbariaux2016} as the backbone for our SBNN training scheme. All experiments use an $\textrm{Adamax}$ optimizer \citep{kingma2014adam} and the Negative Log-Likelihood as loss function. The values of $\gamma$ used to train the SBNNs are reported in \cref{CIFAR_gamma_table}. As previously stated, All experiments were run using 1 GPU from the Google Colaboratory service. All our code and scripts to run our experiments will be made available in a Github repository\footnote{HIDDEN-FOR-SUBMISSION}.
	
	The 2L-MLP and 3L-MLP networks were trained on MNIST with learning rate of $0.01$ and a mini-batch of size 32. The learning rate was decreased by a factor 10 every 15 epochs, over a total of 40 training epochs. No data augmentation was performed on this dataset.
	
	The convolutional topologies were trained on CIFAR-10 and CIFAR-100. At every training epoch, the training set was enlarged using using data augmentation consisting of random rotations up to $\pm15$ degrees and random horizontal flips and crops for every image. Additionally, for CIFAR-100 we used a $\textrm{mixup}$ strategy \cite{zhang2017mixup} with $\alpha=1$ to reduce overfitting. For both datasets, the networks were trained for 300 epochs. On CIFAR-10 we set the learning rate to 0.05, and we decreased it by a factor of 5 every 50 epochs. On CIFAR-100, we set the learning rate to 0.01 and decreased it by a factor of 10 every 120 epochs. In both cases the mini-batch size was set to 200.

%% file: neurips2021.bbl
\begin{thebibliography}{53}
\providecommand{\natexlab}[1]{#1}
\providecommand{\url}[1]{\texttt{#1}}
\expandafter\ifx\csname urlstyle\endcsname\relax
  \providecommand{\doi}[1]{doi: #1}\else
  \providecommand{\doi}{doi: \begingroup \urlstyle{rm}\Url}\fi

\bibitem[Al-Fuqaha et~al.(2015)Al-Fuqaha, Guizani, Mohammadi, Aledhari, and
  Ayyash]{iot_application}
Ala Al-Fuqaha, Mohsen Guizani, Mehdi Mohammadi, Mohammed Aledhari, and Moussa
  Ayyash.
\newblock Internet of things: A survey on enabling technologies, protocols, and
  applications.
\newblock \emph{IEEE communications surveys \& tutorials}, 17\penalty0
  (4):\penalty0 2347--2376, 2015.

\bibitem[Ba and Caruana(2014)]{ba2014}
Jimmy Ba and Rich Caruana.
\newblock Do deep nets really need to be deep?
\newblock In \emph{Advances in Neural Information Processing Systems
  (NeurIPS)}, pages 2654--2662, 2014.

\bibitem[{Bello}(1992)]{Bello1992}
Martin~G. {Bello}.
\newblock Enhanced training algorithms, and integrated training/architecture
  selection for multilayer perceptron networks.
\newblock \emph{IEEE Transactions on Neural Networks}, 3\penalty0 (6):\penalty0
  864--875, 1992.

\bibitem[Bethge et~al.(2019)Bethge, Yang, Bornstein, and
  Meinel]{bethge2019backtosimplicity}
Joseph Bethge, Haojin Yang, Marvin Bornstein, and Christoph Meinel.
\newblock Back to simplicity: How to train accurate bnns from scratch?
\newblock \emph{arXiv preprint arXiv:1906.08637}, 2019.

\bibitem[Canziani et~al.(2016)Canziani, Paszke, and Culurciello]{DNNanalysis}
Alfredo Canziani, Adam Paszke, and Eugenio Culurciello.
\newblock An analysis of deep neural network models for practical applications.
\newblock \emph{arXiv preprint arXiv:1605.07678}, 2016.

\bibitem[Chen et~al.(2017)Chen, Choi, Yu, Han, and Chandraker]{Chen2017}
Guobin Chen, Wongun Choi, Xiang Yu, Tony~X. Han, and Manmohan Chandraker.
\newblock Learning efficient object detection models with knowledge
  distillation.
\newblock In Isabelle Guyon, Ulrike von Luxburg, Samy Bengio, Hanna~M. Wallach,
  Rob Fergus, S.~V.~N. Vishwanathan, and Roman Garnett, editors, \emph{Advances
  in Neural Information Processing Systems (NeurIPS)}, pages 742--751, 2017.

\bibitem[Collobert and Weston(2008)]{natural_language_processing}
Ronan Collobert and Jason Weston.
\newblock A unified architecture for natural language processing: Deep neural
  networks with multitask learning.
\newblock In \emph{Proceedings of the International Conference on Machine
  learning (ICML)}, pages 160--167, 2008.

\bibitem[Courbariaux et~al.(2015)Courbariaux, Bengio, and
  David]{Courbariaux2015}
Matthieu Courbariaux, Yoshua Bengio, and Jean-Pierre David.
\newblock Binaryconnect: Training deep neural networks with binary weights
  during propagations.
\newblock In \emph{Advances in Neural Information Processing Systems
  (NeurIPS)}, pages 3123--3131, 2015.

\bibitem[Denil et~al.(2013)Denil, Shakibi, Dinh, Ranzato, and
  De~Freitas]{denil2013predicting}
Misha Denil, Babak Shakibi, Laurent Dinh, Marc'Aurelio Ranzato, and Nando
  De~Freitas.
\newblock Predicting parameters in deep learning.
\newblock In \emph{Advances in Neural Information Processing Systems
  (NeurIPS)}, pages 2148--2156, 2013.

\bibitem[Ding et~al.(2019)Ding, Chin, Liu, and
  Marculescu]{ding2019regularizing}
Ruizhou Ding, Ting-Wu Chin, Zeye Liu, and Diana Marculescu.
\newblock Regularizing activation distribution for training binarized deep
  networks.
\newblock In \emph{Proceedings of the IEEE/CVF Conference on Computer Vision
  and Pattern Recognition (CVPR)}, pages 11408--11417, 2019.

\bibitem[Evans(2011)]{evans2011}
Dave Evans.
\newblock The internet of things: How the next evolution of the internet is
  changing everything.
\newblock \emph{CISCO white paper}, 1\penalty0 (2011):\penalty0 1--11, 2011.

\bibitem[Faraone et~al.(2017)Faraone, Fraser, Gambardella, Blott, and
  Leong]{faraone2017ternarycompressing}
Julian Faraone, Nicholas Fraser, Giulio Gambardella, Michaela Blott, and
  Philip~HW Leong.
\newblock Compressing low precision deep neural networks using sparsity-induced
  regularization in ternary networks.
\newblock In \emph{International Conference on Neural Information Processing},
  pages 393--404, 2017.

\bibitem[Frankle and Carbin(2019)]{winning_tickets}
Jonathan Frankle and Michael Carbin.
\newblock The lottery ticket hypothesis: Finding sparse, trainable neural
  networks.
\newblock In \emph{International Conference on Learning Representations
  (ICLR)}. OpenReview.net, 2019.

\bibitem[{Global System for Mobile Communications}(2018)]{GSMA_LTE_IoT}
{Global System for Mobile Communications}.
\newblock 3{GPP} low power wide area technologies (white paper).
\newblock Technical report, GSMA, 2018.

\bibitem[Gomez et~al.(2019)Gomez, Zhang, Swersky, Gal, and Hinton]{Gomez2019}
Aidan~N. Gomez, Ivan Zhang, Kevin Swersky, Yarin Gal, and Geoffrey~E. Hinton.
\newblock Learning sparse networks using targeted dropout.
\newblock \emph{CoRR}, abs/1905.13678, 2019.
\newblock URL \url{http://arxiv.org/abs/1905.13678}.

\bibitem[Han et~al.(2015)Han, Pool, Tran, and Dally]{han2015}
Song Han, Jeff Pool, John Tran, and William Dally.
\newblock Learning both weights and connections for efficient neural network.
\newblock In \emph{Advances in Neural Information Processing Systems
  (NeurIPS)}, pages 1135--1143, 2015.

\bibitem[Han et~al.(2016)Han, Mao, and Dally]{compress_prune}
Song Han, Huizi Mao, and William~J Dally.
\newblock Deep compression: Compressing deep neural networks with pruning,
  trained quantization and huffman coding.
\newblock In \emph{International Conference on Learning Representations
  (ICLR)}, 2016.

\bibitem[He et~al.(2016)He, Zhang, Ren, and Sun]{he2016identity}
Kaiming He, Xiangyu Zhang, Shaoqing Ren, and Jian Sun.
\newblock Identity mappings in deep residual networks.
\newblock In \emph{European conference on computer vision}, pages 630--645.
  Springer, 2016.

\bibitem[Hinton et~al.(2012)Hinton, Deng, Yu, Dahl, Mohamed, Jaitly, Senior,
  Vanhoucke, Nguyen, Sainath, et~al.]{speechrecognition}
Geoffrey Hinton, Li~Deng, Dong Yu, George~E Dahl, Abdel-rahman Mohamed, Navdeep
  Jaitly, Andrew Senior, Vincent Vanhoucke, Patrick Nguyen, Tara~N Sainath,
  et~al.
\newblock Deep neural networks for acoustic modeling in speech recognition: The
  shared views of four research groups.
\newblock \emph{IEEE Signal processing magazine}, 29\penalty0 (6):\penalty0
  82--97, 2012.

\bibitem[Hinton et~al.(2015)Hinton, Vinyals, and Dean]{hinton2015}
Geoffrey Hinton, Oriol Vinyals, and Jeff Dean.
\newblock Distilling the knowledge in a neural network.
\newblock \emph{arXiv preprint arXiv:1503.02531}, 2015.

\bibitem[Howard et~al.(2017)Howard, Zhu, Chen, Kalenichenko, Wang, Weyand,
  Andreetto, and Adam]{Howard2017}
Andrew~G Howard, Menglong Zhu, Bo~Chen, Dmitry Kalenichenko, Weijun Wang,
  Tobias Weyand, Marco Andreetto, and Hartwig Adam.
\newblock Mobilenets: Efficient convolutional neural networks for mobile vision
  applications.
\newblock \emph{arXiv preprint arXiv:1704.04861}, 2017.

\bibitem[Hubara et~al.(2016)Hubara, Courbariaux, Soudry, El-Yaniv, and
  Bengio]{Courbariaux2016}
Itay Hubara, Matthieu Courbariaux, Daniel Soudry, Ran El-Yaniv, and Yoshua
  Bengio.
\newblock Binarized neural networks.
\newblock In \emph{Proceedings of the International Conference on Neural
  Information Processing Systems}, pages 4114--4122, 2016.

\bibitem[Hubara et~al.(2017)Hubara, Courbariaux, Soudry, El-Yaniv, and
  Bengio]{courbariaux2017}
Itay Hubara, Matthieu Courbariaux, Daniel Soudry, Ran El-Yaniv, and Yoshua
  Bengio.
\newblock Quantized neural networks: Training neural networks with low
  precision weights and activations.
\newblock \emph{The Journal of Machine Learning Research}, 18\penalty0
  (1):\penalty0 6869--6898, 2017.

\bibitem[Huffman(1952)]{huffman}
David~A Huffman.
\newblock A method for the construction of minimum-redundancy codes.
\newblock \emph{Proceedings of the IRE}, 40\penalty0 (9):\penalty0 1098--1101,
  1952.

\bibitem[Hwang and Sung(2014)]{ternary2014}
Kyuyeon Hwang and Wonyong Sung.
\newblock Fixed-point feedforward deep neural network design using weights+ 1,
  0, and- 1.
\newblock In \emph{IEEE Workshop on Signal Processing Systems (SiPS)}, pages
  1--6, 2014.

\bibitem[Kim and Smaragdis(2015)]{bitwise}
Minje Kim and Paris Smaragdis.
\newblock Bitwise neural networks.
\newblock In \emph{Proceedings of the International Conference on Machine
  Learning JMLR: W\&CP}, volume~37, 2015.

\bibitem[Kingma and Ba(2015)]{kingma2014adam}
Diederik~P Kingma and Jimmy Ba.
\newblock Adam: A method for stochastic optimization.
\newblock In \emph{International Conference on Learning Representations
  (ICLR)}, 2015.

\bibitem[Krizhevsky et~al.()Krizhevsky, Nair, and Hinton]{Cifar10}
Alex Krizhevsky, Vinod Nair, and Geoffrey Hinton.
\newblock Cifar (canadian institute for advanced research).
\newblock Technical report.
\newblock URL \url{http://www.cs.toronto.edu/~kriz/cifar.html}.

\bibitem[Krizhevsky et~al.(2012)Krizhevsky, Sutskever, and Hinton]{alexnet}
Alex Krizhevsky, Ilya Sutskever, and Geoffrey~E Hinton.
\newblock Imagenet classification with deep convolutional neural networks.
\newblock In \emph{Advances in Neural Information Processing Systems
  (NeurIPS)}, pages 1097--1105, 2012.

\bibitem[LeCun and Cortes(2010)]{MNIST}
Yann LeCun and Corinna Cortes.
\newblock {MNIST} handwritten digit database.
\newblock 2010.
\newblock URL \url{http://yann.lecun.com/exdb/mnist/}.

\bibitem[LeCun et~al.(2015)LeCun, Bengio, and Hinton]{bengioDNN}
Yann LeCun, Yoshua Bengio, and Geoffrey Hinton.
\newblock Deep learning.
\newblock \emph{Nature}, 521\penalty0 (7553):\penalty0 436--444, 2015.

\bibitem[Lin et~al.(2020)Lin, Chen, Lin, Cohn, Gan, and Han]{lin2020mcunet}
Ji~Lin, Wei-Ming Chen, Yujun Lin, John Cohn, Chuang Gan, and Song Han.
\newblock Mcunet: Tiny deep learning on iot devices.
\newblock In \emph{Conference on Neural Information Processing Systems
  (NeurIPS)}, 2020.

\bibitem[Liu et~al.(2018)Liu, Wu, Luo, Yang, Liu, and Cheng]{liu2018birealnet}
Zechun Liu, Baoyuan Wu, Wenhan Luo, Xin Yang, Wei Liu, and Kwang-Ting Cheng.
\newblock Bi-real net: Enhancing the performance of 1-bit cnns with improved
  representational capability and advanced training algorithm.
\newblock In \emph{Proceedings of the European Conference on Computer Vision},
  pages 722--737, 2018.

\bibitem[Liu et~al.(2020)Liu, Shen, Savvides, and Cheng]{liu2020reactnet}
Zechun Liu, Zhiqiang Shen, Marios Savvides, and Kwang-Ting Cheng.
\newblock Reactnet: Towards precise binary neural network with generalized
  activation functions.
\newblock In \emph{European Conference on Computer Vision}, pages 143--159.
  Springer, 2020.

\bibitem[Louizos et~al.(2018)Louizos, Welling, and Kingma]{louizos2017}
Christos Louizos, Max Welling, and Diederik~P Kingma.
\newblock Learning sparse neural networks through $l_0$ regularization.
\newblock In \emph{International Conference on Learning Representations
  (ICLR)}, 2018.

\bibitem[Ma et~al.(2018)Ma, Zhang, Zheng, and Sun]{Ma2018}
Ningning Ma, Xiangyu Zhang, Hai{-}Tao Zheng, and Jian Sun.
\newblock Shufflenet {V2:} practical guidelines for efficient {CNN}
  architecture design.
\newblock In Vittorio Ferrari, Martial Hebert, Cristian Sminchisescu, and Yair
  Weiss, editors, \emph{Proocedings of the European Conference on Computer
  Vision {ECCV}}, pages 122--138, 2018.

\bibitem[Marban et~al.(2020)Marban, Becking, Wiedemann, and
  Samek]{marban2020ternarylearning}
Arturo Marban, Daniel Becking, Simon Wiedemann, and Wojciech Samek.
\newblock Learning sparse \& ternary neural networks with entropy-constrained
  trained ternarization (ec2t).
\newblock In \emph{Proceedings of the IEEE/CVF Conference on Computer Vision
  and Pattern Recognition (CVPR)}, pages 722--723, 2020.

\bibitem[McMahan et~al.(2017)McMahan, Moore, Ramage, Hampson, and
  y~Arcas]{mcmahan2017communication}
Brendan McMahan, Eider Moore, Daniel Ramage, Seth Hampson, and Blaise~Aguera
  y~Arcas.
\newblock Communication-efficient learning of deep networks from decentralized
  data.
\newblock In \emph{Artificial Intelligence and Statistics}, pages 1273--1282.
  PMLR, 2017.

\bibitem[{Miraz} et~al.(2015){Miraz}, {Ali}, {Excell}, and
  {Picking}]{Miraz2015}
Mahdi~H. {Miraz}, Maaruf {Ali}, Peter~S. {Excell}, and Rich {Picking}.
\newblock A review on internet of things (iot), internet of everything (ioe)
  and internet of nano things (iont).
\newblock In \emph{2015 Internet Technologies and Applications (ITA)}, pages
  219--224, 2015.

\bibitem[Rastegari et~al.(2016)Rastegari, Ordonez, Redmon, and
  Farhadi]{xnornet}
Mohammad Rastegari, Vicente Ordonez, Joseph Redmon, and Ali Farhadi.
\newblock Xnor-net: Imagenet classification using binary convolutional neural
  networks.
\newblock In \emph{Proceedings of the European Conference on Computer Vision},
  pages 525--542. Springer, 2016.

\bibitem[Recommendation(1988)]{recommendation1988facsimile}
T~Recommendation.
\newblock Facsimile coding schemes and coding control functions for group 4
  facsimile apparatus.
\newblock \emph{International Telecommunication Union, Geneva}, 1988.

\bibitem[Sandler et~al.(2018)Sandler, Howard, Zhu, Zhmoginov, and
  Chen]{Sandler2018}
Mark Sandler, Andrew~G. Howard, Menglong Zhu, Andrey Zhmoginov, and
  Liang{-}Chieh Chen.
\newblock Mobilenetv2: Inverted residuals and linear bottlenecks.
\newblock In \emph{Proceedings of the IEEE/CVF Conference on Computer Vision
  and Pattern Recognition (CVPR), (CVPR) 2018}, pages 4510--4520, 2018.

\bibitem[Simonyan and Zisserman(2015)]{Simonyan2014}
Karen Simonyan and Andrew Zisserman.
\newblock Very deep convolutional networks for large-scale image recognition.
\newblock In \emph{International Conference on Learning Representations
  (ICLR)}, 2015.

\bibitem[Srinivas et~al.(2017)Srinivas, Subramanya, and
  Venkatesh~Babu]{sparse_using_binary}
Suraj Srinivas, Akshayvarun Subramanya, and R~Venkatesh~Babu.
\newblock Training sparse neural networks.
\newblock In \emph{Proceedings of the IEEE/CVF Conference on Computer Vision
  and Pattern Recognition (CVPR)}, pages 138--145, 2017.

\bibitem[Srivastava et~al.(2014)Srivastava, Hinton, Krizhevsky, Sutskever, and
  Salakhutdinov]{srivastava2014}
Nitish Srivastava, Geoffrey Hinton, Alex Krizhevsky, Ilya Sutskever, and Ruslan
  Salakhutdinov.
\newblock Dropout: a simple way to prevent neural networks from overfitting.
\newblock \emph{The Journal of Machine Learning Research}, 15\penalty0
  (1):\penalty0 1929--1958, 2014.

\bibitem[Sze et~al.(2017)Sze, Chen, Yang, and Emer]{sze2017}
Vivienne Sze, Yu-Hsin Chen, Tien-Ju Yang, and Joel~S Emer.
\newblock Efficient processing of deep neural networks: A tutorial and survey.
\newblock \emph{Proceedings of the IEEE}, 105\penalty0 (12):\penalty0
  2295--2329, 2017.

\bibitem[Szegedy et~al.(2015)Szegedy, Liu, Jia, Sermanet, Reed, Anguelov,
  Erhan, Vanhoucke, and Rabinovich]{Szegedy2015}
Christian Szegedy, Wei Liu, Yangqing Jia, Pierre Sermanet, Scott~E. Reed,
  Dragomir Anguelov, Dumitru Erhan, Vincent Vanhoucke, and Andrew Rabinovich.
\newblock Going deeper with convolutions.
\newblock In \emph{Proceedings of the IEEE/CVF Conference on Computer Vision
  and Pattern Recognition (CVPR)}, pages 1--9, 2015.

\bibitem[Tung and Mori(2018)]{tung2018clipq_compression}
Frederick Tung and Greg Mori.
\newblock Clip-q: Deep network compression learning by in-parallel
  pruning-quantization.
\newblock In \emph{Proceedings of the IEEE/CVF Conference on Computer Vision
  and Pattern Recognition (CVPR)}, pages 7873--7882, 2018.

\bibitem[Umuroglu et~al.(2017)Umuroglu, Fraser, Gambardella, Blott, Leong,
  Jahre, and Vissers]{umuroglu2017finn}
Yaman Umuroglu, Nicholas~J Fraser, Giulio Gambardella, Michaela Blott, Philip
  Leong, Magnus Jahre, and Kees Vissers.
\newblock Finn: A framework for fast, scalable binarized neural network
  inference.
\newblock In \emph{Proceedings of the ACM/SIGDA International Symposium on
  Field-Programmable Gate Arrays}, pages 65--74, 2017.

\bibitem[Yang et~al.(2019)Yang, Shen, Xing, Tian, Li, Deng, Huang, and
  Hua]{yang2019quantization}
Jiwei Yang, Xu~Shen, Jun Xing, Xinmei Tian, Houqiang Li, Bing Deng, Jianqiang
  Huang, and Xian-sheng Hua.
\newblock Quantization networks.
\newblock In \emph{Proceedings of the IEEE/CVF Conference on Computer Vision
  and Pattern Recognition (CVPR)}, pages 7308--7316, 2019.

\bibitem[Yao et~al.(2018)Yao, Zhao, Zhang, Hu, Shao, Zhang, Su, and
  Abdelzaher]{yao2018}
Shuochao Yao, Yiran Zhao, Aston Zhang, Shaohan Hu, Huajie Shao, Chao Zhang,
  Lu~Su, and Tarek Abdelzaher.
\newblock Deep learning for the internet of things.
\newblock \emph{Computer}, 51\penalty0 (5):\penalty0 32--41, 2018.

\bibitem[Zhang et~al.(2018{\natexlab{a}})Zhang, Yang, Ye, and Hua]{Zhang2018}
Dongqing Zhang, Jiaolong Yang, Dongqiangzi Ye, and Gang Hua.
\newblock Lq-nets: Learned quantization for highly accurate and compact deep
  neural networks.
\newblock In Vittorio Ferrari, Martial Hebert, Cristian Sminchisescu, and Yair
  Weiss, editors, \emph{Proocedings of the European Conference on Computer
  Vision}, volume 11212, pages 373--390, 2018{\natexlab{a}}.

\bibitem[Zhang et~al.(2018{\natexlab{b}})Zhang, Cisse, Dauphin, and
  Lopez-Paz]{zhang2017mixup}
Hongyi Zhang, Moustapha Cisse, Yann~N Dauphin, and David Lopez-Paz.
\newblock mixup: Beyond empirical risk minimization.
\newblock In \emph{International Conference on Learning Representations
  (ICLR)}, 2018{\natexlab{b}}.

\end{thebibliography}
